\documentclass[manuscript]{acmart}

\AtBeginDocument{%
  }

\acmISBN{978-1-4503-XXXX-X/18/06}

\usepackage{algpseudocode}
\usepackage{algorithm}
\usepackage{bbm}
\usepackage{tabularx}
\newcolumntype{P}[1]{>{\raggedright\arraybackslash}p{#1}}

\usepackage{xspace}
\usepackage{soul}
\PassOptionsToPackage{svgnames}{xcolor}

\definecolor{lightgreen}{HTML}{7fc97f}
\definecolor{lightpurple}{HTML}{984ea3}
\definecolor{lightorange}{HTML}{fdc086}
\definecolor{lightblue}{HTML}{386cb0}

\newcommand{\deparadoxtree}{{\scshape \color{black} De-paradox Tree}\xspace}

\newcommand{\revision}[1]{{\textcolor{black}{#1}}}

\newcommand{\xteng}[1]{{\textcolor{black}{#1}}}
\newcommand{\xtengR}[1]{{\textcolor{black}{#1}}}
\newcommand{\xtengRp}[1]{{\textcolor{black}{#1}}}

\newcommand{\yrv}[1]{{\color{black}{#1}}}
\begin{document}

\title{De-paradox Tree: Breaking Down Simpson's Paradox via A Kernel-Based Partition Algorithm}

% \author{Anonymous authors}

\author{Xian Teng}
\email{xit22@pitt.edu}
\orcid{1234-5678-9012}
\author{Yu-Ru Lin}
\authornote{Corresponding author}
\email{yurulin@pitt.edu}
\orcid{0000-0002-8497-3015}
\affiliation{%
  \institution{University of Pittsburgh}
  \streetaddress{4200 Fifth Ave}
  \city{Pittsburgh}
  \state{Pennsylvania}
  \country{USA}
  \postcode{15260}
}

\begin{abstract}

\yrv{\xtengRp{Real-world observational datasets} and machine learning have revolutionized data-driven decision-making, yet many models rely on empirical associations that may be misleading due to confounding and subgroup heterogeneity. Simpson’s paradox exemplifies this challenge, where aggregated and subgroup-level associations contradict each other, leading to misleading conclusions. Existing methods provide limited support for detecting and interpreting such paradoxical associations, especially for practitioners without deep causal expertise. We introduce \deparadoxtree, an interpretable algorithm designed to uncover hidden subgroup patterns behind paradoxical associations \xtengRp{under assumed causal structures involving confounders and effect heterogeneity.} It employs novel split criteria \xtengRp{and balancing-based procedures to adjust for} confounders and homogenize heterogeneous effects through recursive partitioning. Compared to state-of-the-art methods, \deparadoxtree builds simpler, more interpretable trees, selects relevant covariates, and identifies nested opposite effects while ensuring robust estimation of causal effects \xtengRp{when causally admissible variables are provided}. Our approach addresses the limitations of traditional causal inference and machine learning methods by introducing an interpretable framework \xtengRp{that supports non-expert practitioners while explicitly acknowledging causal assumptions and scope limitations, enabling more reliable and informed decision-making in complex observational data environments.}}

\end{abstract}



\maketitle

\section{Introduction}\label{sec:introduction}

\begin{figure}
    \centering
    \includegraphics[width=\linewidth]{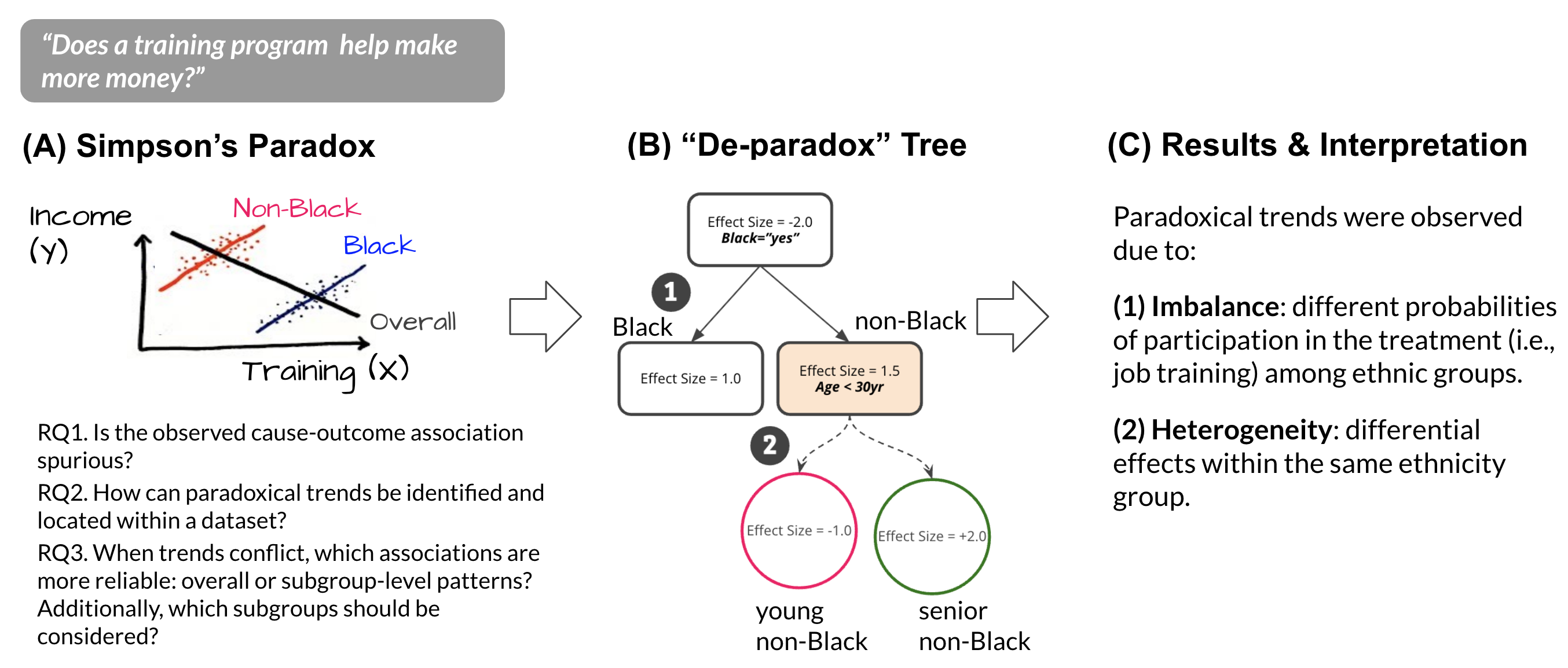}
    \caption[Overview of the proposed \deparadoxtree algorithm.]{
    {This work provides a \deparadoxtree method to address empirical spurious and paradoxical associations that might lead to misleading interpretations of causal effects.}
    {(A) Simpson's paradox, are prevalent in observational studies. E.g., in the study that investigates the effect of a job training program (ref. Section~\ref{sec:introduction}), the cause (i.e., job training program) and outcome (i.e., earnings) can be distorted by a third variable (i.e., ethnicity), leading to a misleading interpretation of the causal effect. }
    {(B) Suppose that a tree produced by \deparadoxtree (hypothetical for illustration) by automatically decomposing cause-outcome associations in data, recursively splitting the data through (1) minimizing the imbalance of pretreatment variables over the treated and control arms, and (2) homogenizing inconsistent effects within subgroups.}
    {(C) The proposed \deparadoxtree helps to explain why paradoxical associations may appear in data. This can be due to (1) {\it imbalance}, where {\it ethnicity} plays as a confounding variable, causing different ethnic groups have different probabilities of participating in job training program, and (2) {\it heterogeneity}, where {\it age} plays as an effect modifier and the job training has different effects within the same ethnic group when decomposing by age.}
    }
    \label{fig:deparadox_tree}
\end{figure}

Decision making based on observational data is ubiquitous in current society. However, typical data analysis practices and machine learning (ML) methods (such as regressions) are designed to capture associations from real data, which often encounter {\it spurious associations} or {\it paradoxical associations} instead of the true {\it causal relationships} \cite{degrave2021ai,joshi2022all,park2021comparison}. \revision{A well-known phenomenon is Simpson's Paradox \cite{kievit2013simpson,lerman2018computational,von2021simpson}, first described by Karl Pearson et al. in 1899 \cite{professor1899genetic} and then named by Edward Simpson in 1951 \cite{simpson1951interpretation}, referring to that the direction of an association between two variables $X$, $Y$ in aggregate may be different after decomposing into subgroups. A popular study concerns whether a government-funded training program helps to increase citizens' income \cite{lalonde1986evaluating,dehejia2002propensity,dehejia1999causal}, seen in Figure~\ref{fig:deparadox_tree} (A). Data shows that program participants earn less money than non-participants.\footnote{\url{https://users.nber.org/~rdehejia/nswdata2.html}} However, when decomposing participants based on ethnicity (i.e., Black and non-Black), program participants earned more than non-participants in both subgroups. The paradox arises because Black people are at a lower income level due to sociodemographic disadvantages, meanwhile more Black joined the program than non-Black driven by their needs for financial improvement. A similar
paradox has been presented in an article of Jeffrey Morris \cite{jeffrey2021israeli} that a high proportion 60\% of patients hospitalized for COVID-19 are vaccinated. But when stratifying the patients into two subgroups based on age (<50yr, >50yr), the vaccine effectiveness against severe illness
is very high (91.8\% and 85.2\%). It is because older people are more likely to be vaccinated, meanwhile they are at higher risk of severe
diseases that require hospitalization. Without knowledge and tools to tackle Simpson's paradox, people might use the misleading overall associations as strong causality evidence to make improper decisions, such as terminating the job training program, or pushing for anti-vaccine movements. More Simpson's examples and the urge to investigate Simpson's paradox for data-driven decision-making have been discussed in computational social science \cite{lerman2018computational}, psychological science \cite{kievit2013simpson}, and health services \cite{obermeyer2019dissecting}.}

\begin{figure}[h]
    \centering
    \includegraphics[width=0.4\linewidth]{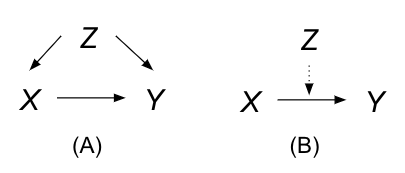}
    \caption[Two causal structures for Simpson's paradox] {\revision{Two causal structures for Simpson's paradox \xtengRp{considered} in this study. (A) A confounder $Z$ influences both cause $X$ and outcome $Y$, causing the marginal and conditional associations diverge. (B) A third variable $Z$ does not confound $X$ and $Y$, but the conditional associations differ \xtengRp{across levels of $Z$ due to effect heterogeneity, and} from each other when conditioning on $Z$. The marginal association is also different from conditional associations. \xtengRp{Other causal structures, such as collider-based structures, are intentionally not shown, as conditioning on colliders may induce selection bias and invalidate causal interpretations in balancing-based analyses---an important consideration when the method is used by non-expert practitioners.}}}
    \label{fig:causal_structure}
\end{figure}

\yrv{
\paragraph{Research gaps.}
Many researchers have studied Simpson's paradox from causal perspectives (see Section~\ref{sec:simpson_paradox} for details). Pearl provides a causal explanation using causal graphs to represent variables and their causal relationships \cite{pearl2003causality}. Hern{\'a}n et al. \cite{hernan2011simpson} reviewed three causal structures behind Simpson's examples. \xtengRp{This study only focuses on two common causal structures behind Simpson's paradox. Additional scenarios discussed by Pearl \cite{pearl2003causality} and Hern{\'a}n et al. \cite{hernan2011simpson} such as collider-based causal structure is out of the scope of this study.}

\begin{enumerate}
    \item \xtengR{{\bf Confounder Imbalance.} Figure~\ref{fig:causal_structure} (A) depicts the most widely discussed causal structures underlying Simpson’s paradox. When a third confounding variable $Z$ influences both $X$ ({\it cause}) and $Y$ ({\it outcome}), Simpson’s paradox arises: the aggregated association is reversed compared to each subgroup association; crucially, the subgroup associations remain consistent with one another.}
    \item \xtengR{{\bf Heterogeneous Effects.} Figure~\ref{fig:causal_structure} (B) illustrates the second case where, even when $X$ and $Y$ are not confounded by $Z$, the conditional associations within the subgroups, $Z=1$ and $Z=0$, exhibit heterogeneous effects. A Simpson's paradox driven by heterogeneous effects arises as subgroup associations differ from one another; some may align with the overall aggregated association, while others oppose it.} 
\end{enumerate}
In many cases, these two factors might co-exist, causing the investigation and interpretability of Simpson's paradox very challenging.} Traditional causal inference methods, such as matching \cite{iacus2012causal,ho2007matching} and weighting \cite{hirano2001estimation,robins1994estimation}, aim to adjust for confounding variables to estimate the true causal effect of $X$ on $Y$. However, these methods primarily operate {\bf at the population level and overlook the existence of subgroups}, where causal relationships may vary. Besides, recent research has emphasized subgroup-level causal inference \cite{kunzel2019metalearners,oprescu2019orthogonal,syrgkanis2019machine,mackey2018orthogonal} and the identification of causally ``interesting'' cohorts within populations---such as individuals who respond most to a treatment (e.g., recovering from a disease after taking a medicine), or whose effects differ significantly from others \cite{nagpal2020interpretable,lakkaraju2017learning,ladhania2020learning,tran2019learning}. Methods for subgroup discovery include decision trees \cite{laber2015tree,athey2016recursive}, causal rule sets \cite{wang2022causal,lakkaraju2017learning}, and decision lists \cite{zhang2015using}. While these techniques improve interpretability, they are often designed for randomized control trials (RCTs) rather than complex observational data settings, where confounding and selection bias present additional challenges (see Section~\ref{sec:causal_subgroup_identification} for more detailed discussion).

Furthermore, many causal machine learning methods lack transparency and interpretability, making them difficult to apply in practice. While decision trees offer some interpretability, existing implementations are often developed for expert users in causal inference---those who understand biases, assumptions, and methodological nuances. \xtengRp{However, a broader audience of practitioners, including data analysts, domain experts, and decision-makers, may lack formal training in causal inference \cite{lerman2018computational}. We refer them as ``non-experts'' of ML and causal inference methodologies in this paper who constitute our {\bf target users}.} Without accessible and interpretable tools, they risk misinterpreting spurious associations as meaningful causal relationships, leading to flawed conclusions and misguided decisions.

Our work focuses on addressing these challenges by developing methods that enhance {\bf causal interpretability} for non-expert users, enabling them to navigate conflicting and potentially misleading associations in empirical data analysis.

\yrv{
\paragraph{The present work.}
This work addresses the challenge of identifying and interpreting Simpson’s paradox by proposing a ``{\it de-paradox}'' method to {\bf detect the presence} of spurious cause-outcome associations, {\bf identify} paradoxical trends, and {\bf provide explanations} for decision-making. The key technical question is: {\it How can a data-driven algorithm automatically identify and locate paradoxical trends, mitigate spurious associations due to confounding and heterogeneity, and enhance interpretability for better decision-making?} Specifically, the proposed method will answer the following three questions:

\begin{itemize}
    \item {\bf RQ 1.} Is the observed cause-outcome association spurious?
    \item {\bf RQ 2.} How can paradoxical trends be identified and located within a dataset?
    \item {\bf RQ 3.} When trends conflict, which associations are more reliable---overall or subgroup-level patterns? Additionally, which subgroups should be considered?
\end{itemize}
}

Let us consider again the study about the effect of the training program on income. Ethnicity acts as a {\it confounder} as it influences both {\it treatment} (i.e., whether or not to participate) and {\it outcome} (i.e., the post-training income) and Black acts as an {\it effect modifier} determining treatment effects. \xtengR{To better understand the origins of Simpson’s paradox in this setting, we focus on these two causal factors---{\it confounder imbalance} and {\it heterogeneous effects} \cite{hernan2011simpson,pearl2003causality,pearl2000simpson}---as illustrated in Figure~\ref{fig:causal_structure}.
\begin{enumerate}
    \item {Confounder imbalance} indicates that subgroups might hold disparate treatment preferences or outcomes due to the presence of confounders. For instance, people in ethnicity subgroups (Black vs non-Black) have different probabilities of participation, thus the distributions of covariates (e.g., ethnicity) among participants (i.e., {\it treated}) and non-participants (i.e., {\it control}) are incomparable. When disaggregating on {\it ethnicity}, subgroups show a reversed positive trend relative to the overall negative.
    \item {Heterogeneous effects} refers to the heterogeneous and even opposite {\it causal effects} exhibited by subsets of people. {Hypothetically imagine a scenario where, in the Black group, senior people benefit from the training program whereas junior people do not, suggesting that these two nested, age-based subgroups hold opposite causal effects.} Failure to identify such disparities can lead to biased cause-outcome analysis.
\end{enumerate}}

This work proposes a \deparadoxtree algorithm\footnote{\url{https://github.com/picsolab/deparadox}} that generates a binary decision tree by automatically decomposing cause-outcome associations in data through (1) {\it balancing} covariate distributions between the treated and the control, and (2) {\it homogenizing} inconsistent effects within balanced subgroups, seen in Figure~\ref{fig:deparadox_tree} (B). Its partition criteria are carefully designed to balance covariates between two treatment arms within a subgroup and to search for effect reversal within partitions. To accomplish the objective (1) {\it balancing}, we employ a distribution distance metric, known as {\it kernel distance} \cite{gretton2006kernel,gretton2012kernel,muandet2017kernel} (ref. Section~\ref{sec:deparadox:balancetree}), as the pivotal criterion for expanding the decision tree. The kernel distance effectively captures the disparities in two multidimensional covariate distributions between the treated and control groups. Through the optimization of kernel distance, we can identify a collection of terminal nodes wherein imbalance are effectively alleviated, denoted as ``balanced'' subgroups (ref. Section~\ref{sec:deparadoxtree:policytree}). For objective (2) {\it homogenizing}, we redefine the concept of a homogeneous causal effect by positing that, within a given subgroup all units should uniformly adopt the same treatment (either 1 or 0). This ensures that the anticipated outcome (e.g., welfare) is maximized. If, however, units within the same subgroup must harbor distinct treatment preferences to optimize the collective outcome, we identify the existence of nested subgroups with opposite effects. Guided by this idea, we cultivate a tree using empirical welfare optimization \cite{sverdrup2020policytree} to identify children nodes characterized by opposite effects (ref. Section~\ref{sec:deparadoxtree:policytree}). 

The main contributions are:
\begin{enumerate}
    \item {\bf Target ``de-paradoxing'' Simpson's paradox through growing a binary decision tree for data partition.} We attempt to answer three practical questions {\bf RQ1 -- RQ3} that are important in understanding paradoxical trends. We also identified two pivotal factors -- {\it confounder imbalance} and {\it heterogeneous effects} -- behind Simpson's paradox. The generated tree, applied to empirical data, has the potential to facilitate the identification of paradoxical and opposing effects within subgroups, presenting paradoxical trends to end users (addressing {\bf RQ2, RQ3}). Spurious cause-outcome associations become apparent when downstream subtrees exhibit reversed associations (addressing {\bf RQ1}).
    
    \item {\bf Novel split criteria to address imbalance and heterogeneity.} It employs kernel distance \cite{gretton2006kernel,gretton2012kernel,muandet2017kernel} (ref. Section~\ref{sec:deparadox:balancetree}) to balance covariate distributions by capturing multidimensional disparities between treated and control groups, that are encoded as kernel mean embeddings \cite{fukumizu2004dimensionality,sriperumbudur2010hilbert} (ref. Section~\ref{sec:deparadox:balancetree}). It redefines homogeneous effects as uniform treatment prefernces within subgroups towards empirical welfare optimization \cite{sverdrup2020policytree}. It adopts policy learning \cite{dudik2014doubly,zhou2023offline,kitagawa2018should,athey2021policy}, a tree-based treatment assignment function (ref. Section~\ref{sec:deparadoxtree:policytree}), to search for subgroups with opposite effects. The proposed criteria offer advantages compared to alternatives regarding flexibility and richness, along with computational efficiency and convergence (ref. Section~\ref{sec:deparadox:balancetree}).
    
    \item {\bf Evaluation on synthetic, hybrid, and real-world datasets, including a novel experimental design to quantitatively evaluate a wide range of plausible conditions.} We simulated datasets with distinct imbalance and heterogeneity patterns using various designs for treatment assignment and outcome generation functions (ref. Section~\ref{sec:results::synthetic}). Results show that \deparadoxtree grows the shallowest trees with the most balanced subgroups (ref. Figure~\ref{fig:compr_f}) and achieves the lowest error rate in predicting treatment preferences (ref. Figure~\ref{fig:compr_g}). We also created a hybrid dataset by intentionally introducing imbalance and heterogeneity to assess the method's ability to detect two injected special subgroups (ref. Section~\ref{sec:results::hybrid}). \deparadoxtree successfully identified the correct features at five nodes and detected both subgroups, while the baseline method incorrectly selected features two or three times and failed to identify the subgroups (ref. Figure~\ref{fig:hybridcompr}). Finally, we applied \deparadoxtree to a real dataset for a qualitative analysis to demonstrate how it addresses the three research questions (ref. Section~\ref{sec:results::realistic}).
    \item \xtengRp{{\bf Targeted at non-experts of ML and causal inference methodologies, designed for interpretability, user-friendliness, and accessibility.}} Tailored for non-experts in causal inference or machine learning, practitioners can easily understand the simple binary tree in Figure~\ref{fig:deparadox_tree} (B). \xtengRp{The method assumes that candidate split variables are causally appropriate (e.g., confounders or effect modifiers), rather than colliders.} Tracing down the square subtree reveals ethnicity subgroups that hold distinct participation probabilities, while circular nodes indicate opposite treatment effects within the same subgroup, as summarized in Figure~\ref{fig:deparadox_tree} (C). These findings encourage decision makers to reconsider the training program's effectiveness for the junior non-Black subgroup \xtengRp{under the stated causal assumptions.}
\end{enumerate}

% Our methodology introduces a novel approach for identifying and mitigating spurious cause-outcome associations through automated data partitioning. By leveraging a decision tree framework, the \deparadoxtree, we recursively partition the feature space to achieve balanced covariate distributions and minimize misleading effects. Utilizing the kernel distance metric as the splitting criterion ensures that the partitions accurately reflect distribution discrepancies between treatment arms. Furthermore, we integrate empirical welfare optimization to learn a treatment assignment policy, maximizing expected outcomes by identifying subgroups that benefit from opposing treatments. This method effectively enhances decision-making by distinguishing reliable associations and uncovering heterogeneous treatment effects.
\section{Related Work}\label{cha:relatedwork}

\subsection{Simpson's Paradox Explanation and Detection}\label{sec:simpson_paradox}
\revision{Simpson's paradox is a widely discussed phenomenon whereby the association between a pair of variables $X, Y$ reverses upon conditioning of a third variable $Z$ in a variety of domains \cite{lerman2018computational,kievit2013simpson,alipourfard2021dogr,alipourfard2018can,alipourfard2018using,charig1986comparison,julious1994confounding,huang2015challenges}. Pearl stated that people find it paradoxical as human intuition is governed by causal calculus \cite{pearl2014comment}. In recent years, researches have explained Simpson's paradox using causal terms \cite{lindley1981role,hernan2011simpson,pearl2000simpson}. For example, Hern{\'a}n et al. \cite{hernan2011simpson} reviewed three causal structures including collider, confounder, and effect-modifier, behind Simpson's examples. By explicitly outlining the causal structure of the data, the paradox can be understood and resolved. Pearl developed a theory of causal graphs, which involves a graphical condition called ``back-door'', to identify the class of causal diagrams for which paradoxical observation is realizable \cite{pearl2014probabilistic}. Although previous work links Simpson's paradox to causal explanations, detecting methods of the paradox still remains to be developed in practice.}

\revision{Recently, a few works have proposed automated methods to detect Simpson's paradox. For instance, Xu et al. \cite{xu2018detecting} used categorical variables to partition data into groups and to find association reversal between the coefficient correlations. Alipourfard et al. \cite{alipourfard2018can, alipourfard2018using} present a statistical method to identify Simpson's paradox in data by comparing statistical trends in the aggregate data to those in the disaggregated subgroups. However, the method in \cite{xu2018detecting} requires predefined subgroups to detect Simpson's paradox. If relevant subgroups are not identified beforehand, some paradoxical cases may go undetected. Besides, the above studies only detect Simpson's paradox caused by a single variable and fail to discover those caused by multiple variables. A method called DoGR is proposed in a follow-up study \cite{alipourfard2021dogr}, which simultaneously partitions data into overlapping clusters and models behaviors within them using regressions. This approach assumes that the data within each cluster follows a Gaussian distribution on latent confounders, which may not hold true for all datasets. Secondly, overlapping clusters can capture complex data structures, but they may also introduce redundancy and complicate the interpretation of results. Wang et al. \cite{wang2023learning} proposed a neural network model driven by a loss function called Multi-group Pearson Correlation Coefficient. It's capable of handling both discrete and continuous variables in paradox detection and unifies multiple variants of Simpson's paradox.} \xtengR{However, these methods are purely statistical---they focus on trend reversals without considering underlying causal structures or identifying true causal relationships. As a result, paradoxes detected using these approaches may still be misleading without causal inference techniques.} \yrv{{\bf Our novel causal approach goes beyond statistical trends for improved paradox detection.} Our method overcomes these limitations in three key aspects. First, it accommodates a wide range of variable types while ensuring robust paradox detection. Second, subgroup discovery is fully automated rather than predefined, making the results both data-driven and interpretable. Third, by systematically minimizing confounding biases such as imbalance and heterogeneity, our method enhances causal interpretability and reduces the risk of associational misinterpretation.
}

% \revision{Compared to those methods, our method has a couple of advantages. First, it supports multiple and various types of variables. Second, the discovery of subgroups is automated (rather than predefined) as well as self-explainable (rather than implicit). Third, our study establishes causal relationships through minimizing factors (such as imbalance and heterogeneity) that might cause any associational misinterpretation.}

\subsection{\revision{Causal Inference: Framework and Assumptions}}\label{sec:potential_outcomes_framework}

{Randomized controlled trials (RCTs) are the ``gold standard'' for causal inference, where units (or subjects) are randomly assigned to receive the treatment or not. But in reality they are often unethical, impractical, or untimely \cite{guo2021vaine,guo2023causalvis}, hence causal inference based on observational data has been extensively utilized in many domains \cite{cookson2021equity, clark2015big, gerring2005causation}. RCTs and observational studies are different. RCTs ensure comparable characteristics, {\it balanced}, between treatment groups through randomization, while observational studies may have distinct feature distributions ({\it imbalanced}), making it possible that outcome difference might eventually trace back to factors other than treatment.}

The Potential Outcomes Framework, also called the Rubin Causal Model (RCM) \cite{imbens2010rubin}, is a theoretical framework for causal inference in both observational and experimental studies. Consider again the job training program example in the Introduction, it involves defining two potential outcomes for each person $i$---the potential outcome had they participated in the program $Y_i(1)$, and the outcome had they not $Y_i(0)$. By comparing outcome differences across subjects, the average treatment effect is estimated $\tau=\mathrm{mean}_{i}[Y_i(1)-Y_i(0)]$. The pair of $(Y_i(1), Y_i(0))$ could never be observed at the same time in real life, thus several causal assumptions are critical for making valid causal inferences in observational data. The foremost one is ``{\it no unmeasured confounder}'', any potential confounding factors that could distort the estimated treatment effect are either measured and adjusted for, or simply do not exist. The second one is {\it overlap} or {\it positivity} \cite{heckman1997matching} assuming that participants could have chosen not to attend the program and vice versa with a non-zero probability. Then we have the {\it stable unit treatment value assumption}, SUTVA \cite{vanderweele2013causal,cole2009consistency}, which assumes that subjects maintain independent without intervining each other's treatment decisions and outcomes, and there are no hidden variations of treatment that might lead to distinct outcomes. Additionally, the {\it unconfoundedness} assumption \cite{rosenbaum1983central} requires that the values of the two {potential} outcomes are determined in a manner conditionally irrelevant what treatment to receive. Our work is designed in the context of potential outcome framework on the basis of above assumptions, allowing users to investigate spurious or paradoxical association in observational studies.

\subsection{Confounding and Balancing}\label{sec:confounding_and_balancing}

Empirical cause-outcome associations often deviate from true causal effects due to confounding bias, where certain pre-treatment covariates lack balance across treatment groups. Popular balancing strategies in causal inference literature, such as matching \cite{iacus2012causal,ho2007matching}, stratification \cite{cochran1968effectiveness,rosenbaum1984reducing}, and weighting \cite{hirano2001estimation,robins1994estimation}, have been proposed to address this issue. Matching techniques seek a subpopulation where confounders can be balanced by pairing the closest data points and discarding the unmatched ones. Weighting techniques assign weights to data points to create a synthetic population with balanced confounders. Unlike our work, these methods do not incorporate the concept of subgroups or explore how associational patterns are reflected across them. Stratification methods, while sharing similarities with our study by dividing data points into subsets, also have drawbacks. Raw stratification \cite{iacus2012causal} results in an exponentially growing number of subsets as the number of covariates increases. Propensity score stratification \cite{rosenbaum1983central,rosenbaum1984reducing} is susceptible to model misspecification and lacks interpretability in the original multidimensional feature space. In contrast, our subgroup discovery algorithm addresses these issues. It operates within the original space for interpretability without the need for parametric modeling. Furthermore, our algorithm enhances covariate balance by selectively splitting a few key variables, preventing an exponential increase in partition size as the variable count grows.

The most relevant approaches to this study are covariate-balancing weighting \cite{imai2014covariate,zhao2016covariate,tan2020regularized}, which employs constrained optimization to learn weights for covariate balance between treatment groups. Theoretically speaking, researchers are free to control all powers of each of the covariates, as well as all orders of their interactions, referred as ``parametric covariate balancing.'' But it could become infeasible to balance everything well when the set of interactions or moments grow exponentially in the number of covariates. Some researchers have instead proposed ``nonparametric covariate balancing,'' also called ``kernel balancing'' \cite{ben2021balancing,wong2018kernel,kim2022kernel,hazlettkernel}, which controls the covariate functional balance over a Reproducing-Kernel Hilbert Space (RKHS) \cite{gretton2013introduction}. Kernel balancing is able to balance a richer class of functions of covariates, encompassing interactions and nonlinearities with minimal user intervention. By confining functions within the unit ball of an RKHS, the computational process is streamlined through the ``kernel trick,'' reformulating optimization problems into inner products over finite-dimensional vectors. Drawing inspiration from these approaches, our work utilizes ``kernel distance'' \cite{gretton2006kernel,gretton2012kernel,muandet2017kernel} as a splitting criterion, optimizing it over treated and control groups to identify causal balanced subgroups. Compared to common balance measures (e.g., the overlapping coefficient, the Kolmogorov-Smirnov distance), the kernel distance is proved to be a better metric for the divergence of two distributions as it leverages the benefits of kernel tricks and RKHS flexibility \cite{zhu2018kernel}.

\xteng{Our work proposes a formal, mathematical framework for Simpson's paradox detection that distinguishes it from existing approaches. Unlike prior kernel-based methods, which primarily focus on balancing covariates, our approach provides a rigorous operational definition of Simpson's paradox in terms of causal structures. This mathematical formulation allows for the systematic identification of paradoxical trends as well as subgroup patterns, rather than merely operating at the population level. Besides, while previous methods separately apply balancing techniques and causal effect estimation, our approach integrates causal modeling with tree partitioning, offering a holistic solution for understanding and mitigating Simpson's paradox in observational data.}

\subsection{Causal Subgroup Identification}\label{sec:causal_subgroup_identification}

Many existing works focus on subgroup discovery in the context of causal inference \cite{lemmerich2016fast,zhang2015using,nagpal2020interpretable,lakkaraju2017learning,laber2015tree,athey2016recursive,tran2019learning,wang2022causal,lakkaraju2017learning,seibold2016model,zeileis2008model,su2009subgroup,loh2002regression,gutierrez2017causal,ciampi1988recpam}. Their primary objective is to identify clusters of entities that possess certain characteristics deemed ``interesting'' \cite{lemmerich2016fast}, manifested in distinct ways, such as amplified or diminished treatment effects \cite{zhang2015using,wang2022causal,nagpal2020interpretable}, cost-effective, interpretable, and actionable treatment regimes \cite{lakkaraju2017learning,laber2015tree}, or differing treatment effects from others as much as possible \cite{athey2016recursive,ladhania2020learning,tran2019learning}. Among them, interpretability is a common concern, leading to the utilization of recursive partition methods such as decision trees \cite{laber2015tree,athey2016recursive,tran2019learning}, causal rule sets \cite{wang2022causal,nagpal2020interpretable}, and if-then-else rule sequences \cite{lakkaraju2017learning}. Popular partitioning criteria include highest parameter instability \cite{seibold2016model,zeileis2008model}, statistical tests \cite{su2009subgroup,loh2002regression,gutierrez2017causal}, likelihood ratio statistics \cite{ciampi1988recpam}, and penalizing splits with high variance \cite{athey2016recursive}. Additionally, there are concurrent endeavors in policy learning and optimization \cite{athey2021policy,zhou2023offline,dudik2011doubly}, aiming to determine the optimal ``policy'' of assigning treatments based on historical data to maximize the collective welfare (i.e., outcomes). The most relevant method to this work is ``policy tree'' \cite{sverdrup2020policytree}, employing a decision tree as the treatment assignment function with a predefined depth to partition the population into subsets with distinct recommended treatments. Our work borrows policy tree as a way to detect effect reversal. If units within the same subgroup shall take the opposite treatments to optimize the collective outcomes, we identify the existence of heterogeneous effects. 

% While policy learning shares a common objective of individualized treatment assignments with the previously mentioned subgroup discovery studies, their split criteria diverge significantly. The former strives to maximize a notion of average social welfare \cite{kitagawa2018should}, while the latter methods pursue objectives such as squared-error loss or significance-based criteria.

In summary, the previous studies on subgroup discovery focused on personalized treatment assignments based on effects and costs. In contrast, our objective is to examine Simpson's paradox, specifically exploring how cause-outcome associations manifest across subgroups while accounting for confounding bias and heterogeneous subgroup effects. \xteng{Our method goes beyond the aforementioned tree-growing methods \cite{laber2015tree,athey2016recursive,tran2019learning} by incorporating a welfare-based optimization framework that directly links subgroup partitioning to outcome optimization. This policy-learning component ensures that detected subgroups are not just statistically distinct but also actionable for decision-making.}

\section{Problem Settings}\label{cha:problemsetting}
\subsection{Notations}\label{sec:notations}
Suppose we have a observational dataset consisting of i.i.d triplets $\mathcal{D} \triangleq \{(X_i,Y_i,\mathbf{Z}_i), i=1,...,N\}$ , where $X_i$ is a binary treatment for unit $i$ (either 0 or 1, \revision{indicating the unit receiving treatment or not}), $Y_i$ is its outcome, and $\mathbf{Z} \in \mathcal{Z} $ is its $K$-dimensional feature vector in a space $\mathcal{Z} \subset \mathbb{R}^K$. To depict relationships between elements $X, Y, \mathbf{Z}$, we use $\rightarrow$ for causal relations \cite{pearl1995causal} and $\leftrightarrow$ for associations. Hence, $\mathbf{Z}\rightarrow X$ signifies treatment propensity \cite{rosenbaum1983central}, reflecting the inclination of $\mathbf{Z}$-defined subjects to receive specific treatment, $\mathbf{Z}\rightarrow Y$ captures a subject's inherent outcome tendency without active treatment, $X\rightarrow Y$ denotes the true causal effect, and $X\leftrightarrow Y$ indicates empirical cause-outcome association. As shown in Figure~\ref{fig:causal_structure} (A), treatment imbalance means that, as $\mathbf{Z} \not\!\perp\!\!\!\perp X$ since $\mathbf{Z}$ affects $X$, the participartion probability $X|\mathbf{Z}$ and $X|\mathbf{Z}^{\prime}$ are not the same. Covariate distributions, $\mathbf{Z}|X=0$ and $\mathbf{Z}|X=1$, are not comparable accordingly, which is referred as {\it covariate imbalance}. Therefore, the cause-outcome associations are distorted as outcome difference of the treated against the control might trace back to difference in $\mathbf{Z}$ beyond $X$. Shown in Figure~\ref{fig:causal_structure} (B), heterogeneous effects suggests that an overall causal effect does not generalize to subgroups, which might hold different or opposite trends. Please see Table~\ref{tab:symbols:deparadox} for a list of symbols.

\subsection{Overview of the \deparadoxtree}\label{sec:overview_deparadoxtree}

\yrv{
Figure~\ref{fig:deparadox_tree} (B) provides a schematic illustration of the \deparadoxtree algorithm. The method automatically partitions data to minimize imbalance and homogenize effects, reducing spurious cause-outcome associations.
The algorithm consists of two stages:
\begin{itemize}
    \item {\bf Stage 1: Causal Balance Tree} (Figure~\ref{fig:deparadox_tree} (B) square nodes): This stage partitions data to balance covariate distributions at each split. In Figure~\ref{fig:deparadox_tree} (B), square nodes with solid arrows represent this process. Ethnicity is identified as the optimal splitting feature, resulting in two balanced subgroups: {Black} and {non-Black}. The effect sizes in these subgroups are reversed compared to the overall trend.
    \item {\bf Stage 2: Opposite-Effects Policy Tree} (Figure~\ref{fig:deparadox_tree} (B) circular nodes): After forming balanced subgroups, this stage analyzes whether nested subgroups exhibit different treatment preferences. Circular nodes with dashed arrows in Figure~\ref{fig:deparadox_tree} (B) represent this process. The {non-Black} subgroup further splits based on age, forming {young non-Black} and {senior non-Black} nodes. The young subgroup demonstrates a negative effect size, whereas the senior subgroup exhibits a positive effect size, likely benefiting from the training program.
\end{itemize}
The example demonstrates how our proposed method directly addresses the three RQs by illustrating the algorithm’s ability to detect paradoxical trends, identify nested subgroup effects, and explain treatment heterogeneity. 
}

\yrv{
\begin{itemize}
    \item {\bf RQ 1.} {\it Is the observed cause-outcome association spurious?}
{\bf Answer:} Yes, the overall association is misleading due to confounding and heterogeneity. Ethnicity acts as a confounder, introducing imbalance. Once controlled for, different trends emerge at the subgroup level. Age introduces heterogeneity in the estimated causal effects---when stratified by age, two distinct subgroups with opposing responses to the training program become apparent.
    \item {\bf RQ 2.} {\it How can paradoxical trends be identified and located within a dataset?}
{\bf Answer:} The proposed method systematically identifies subgroups that minimize imbalance and heterogeneity. By following the tree structure, trend reversals become apparent. For instance, in {\bf Stage 1} (square nodes), the Black and non-Black subgroups exhibit reversed trends relative to their parent node. In {\bf Stage 2} (circular nodes), the non-Black young subgroup displays a negative trend, in contrast to the positive trend of its parent node.
    \item {\bf RQ 3.} {\it When trends conflict, which associations are more reliable---overall or subgroup-level patterns? Which subgroups should be considered?}
{\bf Answer:} The method identifies three key subgroups: Black, non-Black young, and non-Black senior. Subgroup-level associations are more reliable than the overall trend because the tree-based approach systematically accounts for confounding and heterogeneity. By partitioning the data into homogeneous subgroups, the method ensures that observed associations better reflect true causal effects rather than misleading aggregated patterns.
\end{itemize}
}

We refer the first stage as growing a {\it casual balance tree} (ref. Section~\ref{sec:deparadox:balancetree}), while the second stage as growing an {\it opposite-effects policy tree} (ref. Section~\ref{sec:deparadoxtree:policytree}).
\xtengR{The rationale for placing imbalance first is causal: confounder imbalance produces spurious associations that can obscure or distort heterogeneity. By prioritizing its removal in the first stage, we ensure that any heterogeneous effects identified in the second stage can be interpreted as genuine effect modification rather than residual bias. The sequential design also brings flexibility in tree building: When imbalance is absent, the Stage 1 causal balance tree terminates immediately, and the Stage 2 opposite-effects policy tree starts to split at this root node; similarly, when heterogeneity is absent, only a causal balance tree would grow, without producing any nested policy trees at the terminal nodes. When both factors are absent, the root node remains intact. The flexible design of \deparadoxtree is applicable to both single-factor-driven paradoxes (either heterogeneity or imbalance), but also dual-factor-driven paradoxes.}

\xtengRp{Applying the proposed method to real-world datasets requires domain knowledge to accurately infer causal structures. The method is specifically designed for scenarios that align with the two causal structures shown in Figure~\ref{fig:causal_structure}, but it is not suitable for cases where {\it colliders} distort associations \cite{hernan2011simpson}.} \xtengR{Importantly, the presence of Simpson’s paradox is not a prerequisite---some datasets may already have well-balanced covariates and homogeneous effect sizes. In such cases, the root node itself serves as the final \deparadoxtree, making the overall association a reliable reference for decision-making. When paradoxical trends are present, however, the method is capable of detecting trend reversals and isolating heterogeneous treatment effects, providing deeper insight into subgroup behaviors.}

\begin{table}[h]
\centering
\caption{A table of math symbols used in the proposed \deparadoxtree algorithm.}
\begin{tabular}{r l}
{\bf Symbols}  & {\bf Explanations} \\
\toprule
$\mathcal{D} \triangleq \{X_i,Y_i,\mathbf{Z}_i\}$ & The observational data consisting of i.i.d. triplets $\{X_i,Y_i,\mathbf{Z}_i\}$ \\
$p, q$ & The feature distributions for the treated/control arms\\
$D[p, q], D_{\mathcal{H}_1}[p,q]$ & A generic distance as well as kernel distance of two distributions $p,q$ \\
$\mathcal{F}$ & A function class \\
$\mathcal{H}_1$ & The unit ball of a reproducing kernel Hilbert space (RKHS) \\
$\langle \cdot, \cdot \rangle_{\mathcal{H}}$ & Inner product of a RKHS \\ 
$\kappa(\cdot, \cdot)$ & Kernel operation \\
$\phi: \mathcal{Z} \longrightarrow \mathcal{H}$ & A mapping function from feature space to an alternative space \\
$\mu_p, \mu_q$ & Kernel mean embeddings of $p,q$ in a RKHS \\
$\pi:\mathcal{Z}\longrightarrow\{0,1\}$ & A policy mapping features into treatments \\
$W(\pi), \widehat{W}(\pi)$ & Expected welfare given a policy and its empirical estimator \\
% $R(\pi)$ & Regret of a policy \\
${T},T_l,T_r$ & A decision tree or a node, the left and right subtree \\
$s\triangleq\{k, z_k\}$ & A split including a feature $k$ and its cutoff point $z_k$ \\
$\widehat{e}$ & Propensity score model \\
$\widehat{m}$ & Response model \\
% $\widehat{\Gamma}_i$ & The doubly robust estimator vector for unit $i$ \\
\bottomrule
\end{tabular}
\label{tab:symbols:deparadox}
\end{table}
\section{Methods}\label{sec:deparadox:methods}

\subsection{Stage 1: Growing A Causal Balance Tree}\label{sec:deparadox:balancetree}
\begin{figure}[h]
    \centering
    \includegraphics[width=0.6\linewidth]{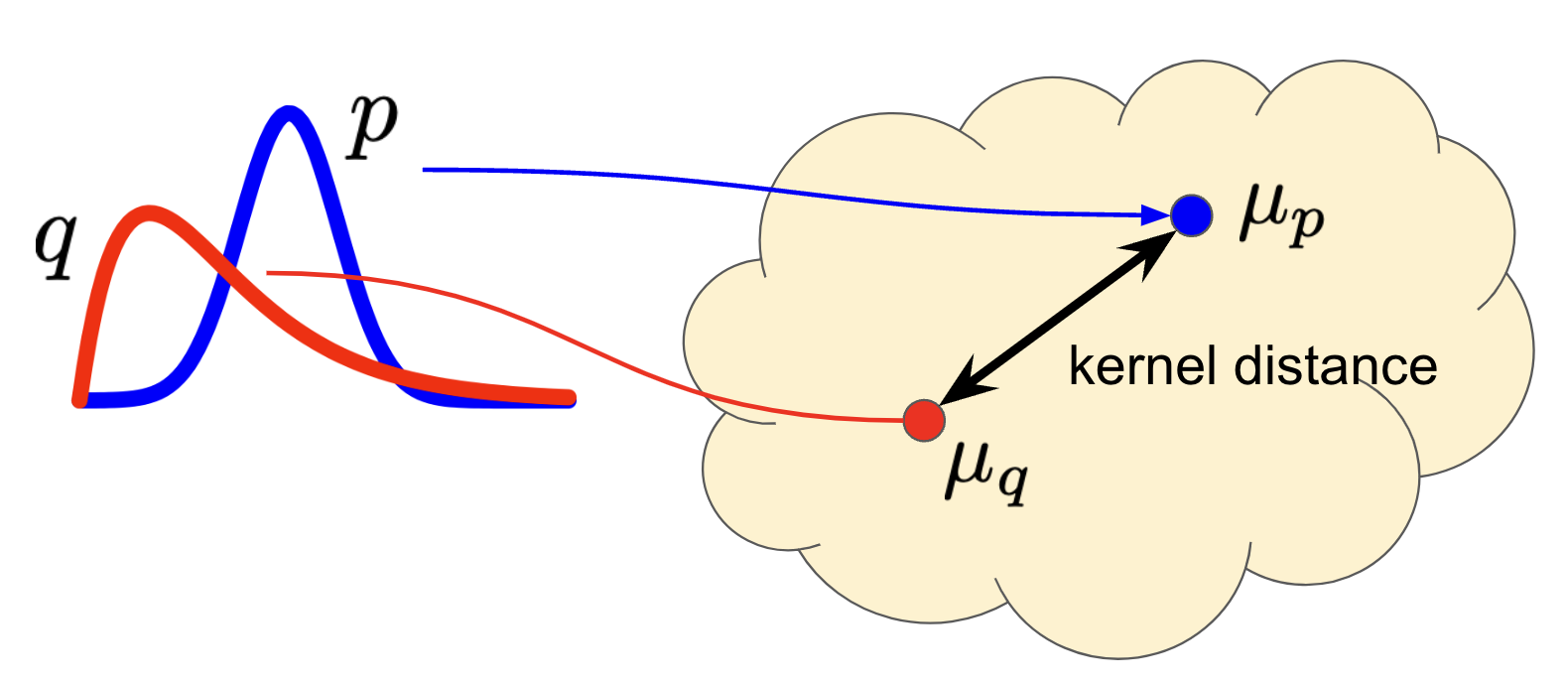}
    \caption[Kernel mean embeddings and kernel distance.]{Kernel mean embeddings and kernel distance. Probability distributions are uniquely mapped into data points---i.e., kernel mean embeddings---in a reproducing kernel Hilbert space (RKHS) through an expectation operation, so that a kernel distance of two distributions could be computed via the distance of these two data embeddings.}
    \label{fig:kernel_embeddings}
\end{figure}
\revision{The first stage of causal balance tree aims to mitigate treatment imbalance through recursive splitting to search for balanced subgroups. At each step of splitting, it chooses the best split that leads to the greatest decrease in a distance of two multivariate distributions, $D[p(\mathbf{Z}|X=1), q(\mathbf{Z}|X=0)]$, abbreviated as $D[p,q]$. In \deparadoxtree, this distance is measured using kernel distance \cite{gretton2006kernel,gretton2012kernel,muandet2017kernel}, a metric derived from the framework of kernel mean embeddings \cite{muandet2017kernel,fukumizu2004dimensionality,sriperumbudur2010hilbert}. As shown in Figure~\ref{fig:kernel_embeddings}, the core idea of kernel mean embeddings is to map a distribution $p$ from the initial feature space $\mathcal{Z}$ to an unique vector $\mu_p$ in a specialized Hilbert space $\mathcal{H}$ \cite{gretton2013introduction}. This transformation allows the distance between distributions to be computed as the squared distance between their corresponding vectors: $D[p, q]=\|\mu_p - \mu_q\|^2$.}

\subsubsection{Kernel Distance} Kernel mean embedding has recently emerged as a powerful tool for machine learning and statistical inference \cite{muandet2017kernel}. This framework owes its success to a positive definite function commonly known as the {\it kernel functions}. Let $\kappa: \mathcal{Z}\times\mathcal{Z}\rightarrow\mathbb{R}$ be a symmetric and positive definite kernel function that defines a {\it feature map} $\mathbf{z} \mapsto \phi(\mathbf{z}) := \kappa(\mathbf{z},\cdot)\in \mathcal{H}$ satisfying $ \kappa(\mathbf{z},\mathbf{z}^{\prime}) = \langle \phi(\mathbf{z}),\phi(\mathbf{z}^{\prime})\rangle_{\mathcal{H}}$. The space $\mathcal{H}$ is a reproducing kernel Hilbert space where inner product $\langle\cdot,\cdot\rangle_{\mathcal{H}}$ is defined and $\kappa$ is its {\it reproducing kernel}. The RKHS $\mathcal{H}$ is fully characterized by the reproducing kernel $\kappa$, in a sense that $\kappa(\cdot, \cdot)$ (or $\phi$) maps each point $\mathbf{z}\in\mathcal{Z}$ to an element $\kappa(\cdot, \mathbf{z})$ (or $\phi(\mathbf{z})$) in $\mathcal{H}$, and the kernel of two elements in $\mathcal{Z}$ is equal to the inner product of their feature map functions in $\mathcal{H}$. Refer to Appendix~\ref{appendix:kernels} and ~\ref{appendix:rkhs} for more information about kernel method and RKHS. Kernel mean embedding is to extend the feature map $\phi$ from individual data points to the space of probability measures $\mathcal{P}$ by representing each distribution $p$ as a mean function:
\begin{equation}
    \mu_p = \phi(p) := \int_{\mathcal{Z}}\phi(\mathbf{Z})dp(\mathbf{Z}).
\end{equation}
In this case we essentially transform the distribution $p$ to an element $\mu_p\in\mathcal{H}$ in the RHKS space $\mathcal{H}$, so that the kernel mean representation $\mu_p$ captures all information about the distribution $p$ \cite{fukumizu2004dimensionality,sriperumbudur2010hilbert}. Ideally, the mean mapping $p\mapsto\mu_p$ should be {\it injective}, that is, $\|\mu_p - \mu_q\|_{\mathcal{H}}=0$ if and only if $p=q$. 
In other words, it should create a one-to-one mapping between distributions and the embeddings so that there is no information loss during the transformation. 

Gretton et al. \cite{gretton2012kernel, gretton2006kernel} establish a condition on the RKHS to satisfy this condition. Their works considered $\mathcal{H}$ to be a special RKHS where all elements in this space is limited by an unit ball, i.e., $\mathcal{H}_1=\{f:\|f\|_{\mathcal{H}}\leq 1\}$, and the distance of $p,q$ is referred as {\it kernel distance} \cite{gretton2006kernel,gretton2012kernel,muandet2017kernel}: 
\begin{equation}
    D_{\mathcal{H}_1}[p,q] = \|\mu_{p} - \mu_{q}\|_{\mathcal{H}_1}^2.
    \label{equation:kernel_distance}
\end{equation}
Although Equation~(\ref{equation:kernel_distance}) involves two kernel mean embeddings in $\mathcal{H}_1$: $\mu_p, \mu_q$, in practices we do not need to explicitly compute them. Instead, the right-hand side can be expanded as:
\begin{equation}
    \|\mu_{p} - \mu_{q}\|_{\mathcal{H}_1}^2 = \mathrm{E}_{\mathbf{Z}\sim p,\mathbf{Z}\sim p}\big[\kappa(\mathbf{Z},\mathbf{Z})\big] + \mathrm{E}_{\mathbf{Z}^{\prime}\sim q,\mathbf{Z}^{\prime}\sim q}\big[\kappa(\mathbf{Z}^{\prime},\mathbf{Z}^{\prime})\big] -2 \mathrm{E}_{\mathbf{Z}\sim p,\mathbf{Z}^{\prime}\sim q}\big[\kappa(\mathbf{Z},\mathbf{Z}^{\prime})\big],
\end{equation}
where $\kappa$ provides a convenient solution by simply computing kernel values in the original space (the so-called ``kernel tricks''). \revision{In real-world scenarios, we typically have little knowledge about these two distributions of $p,q$. Suppose there are $n_1$ feature vectors $\mathbf{Z}_i$ from the treated data samples (representing $p$), and $n_0$ vectors $\mathrm{Z}_j$ from the control samples (representing $q$),} we can compute an unbiased empirical estimate through those observed feature vectors \cite{gretton2012kernel}:
\begin{equation}
    \widehat{D}_{\mathcal{H}_1}[p,q] = \frac{1}{n_1(n_1-1)}\sum_{i=1}^{n_1}\sum_{j\neq i}^{n_1}\kappa(\mathbf{Z}_i,\mathbf{Z}_j) + \frac{1}{n_0(n_0-1)}\sum_{i=1}^{n_0}\sum_{j\neq i}^{n_0}\kappa(\mathbf{Z}^{\prime}_i,\mathbf{Z}^{\prime}_j) - \frac{2}{n_1n_0}\sum_{i=1}^{n_1}\sum_{j=1}^{n_0}\kappa(\mathbf{Z}_i,\mathbf{Z}^{\prime}_j).
    \label{equation:kernel_distance_estimator}
\end{equation}
The choice of kernel distance offers advantages compared to alternative covariate imbalance (or distance) metrics that have been used in causal literature \cite{austin2015moving,austin2009balance}. (1) {\it Flexibility and richness}: Kernel distance depends solely on the kernel and allows for flexible kernel definition on arbitrary domains. Prior works \cite{fukumizu2004dimensionality,sriperumbudur2010hilbert} have proven that the space $\mathcal{H}_1$ used in kernel distance ensures an ``injective'' one-to-one mapping and uniquely encode distributions into vectors in $\mathcal{H}_1$. Other metrics like standardized differences of means do not satisfy this injective mapping criterion. (2) {\it Computational efficiency and convergence} \cite{gretton2012kernel}: \revision{Kernel distance is computationally efficient, with a time complexity of $O((n_1 + n_0)^2)$ (suppose $n_1$ data points from $p$ and $n_0$ data points from $q$).} The empirical estimate converges rapidly, and this rate is independent of the dimension $K$ of the space $\mathbb{R}^K$. In contrast, metrics like Kullback-Leibler divergence may exhibit arbitrarily slow convergence rates depending on the distributions. Appendix~\ref{appendix:kernel_distance} provides more information about the good properties of kernel mean embedding and kernel distance. Following \cite{wong2018kernel}, we have used the radial basis function (RBF) kernel, as it has been demonstrated to satisfy all derivative constraints simultaneously and ensures a strict definition for balance.

\begin{algorithm}[h]
\caption{\yrv{
$\mathrm{\bf CausalBalanceTree}$: Balance pretreatment covariate distributions at each split.}
}
\label{code:causalbalance}
\begin{algorithmic}[1]
    \Require Observational data $\mathcal{D}$, tree expansion stopping criteria (e.g., maximum depth, minimum leaf size, etc), globally sort $N$ samples along $K$ features and cache the indices, \xtengRp{pre-compute a kernel matrix $\kappa$ of size $N\times N$, which constitutes the dominant computational and memory cost of the method and is reused in every split step.}
    \State $T^1 \longleftarrow \emptyset$
    \If{not meeting the stopping criteria,}
    \State Create a new node, compute its kernel distance score $\widehat{D}_{\mathcal{H}_1}[p,q]$ according to Equation~(\ref{equation:kernel_distance_estimator})
    \State Find the best split $s^{\ast}$ through Equation~(\ref{equation:bestsplit}), and split $\mathcal{D}$ into $\mathcal{D}_l, \mathcal{D}_r$ according to $s^{\ast}$
    \State \revision{Learn a left tree ${T}^1_l \longleftarrow \mathrm{CausalBalanceTree}(\mathcal{D}_l)$}
    \State \revision{Learn a right tree ${T}^1_r \longleftarrow \mathrm{CausalBalanceTree(\mathcal{D}_r)}$}
    \State ${T}^1 \longleftarrow \{s^{\ast},({T}^1_l, {T}^1_r)\}$
    \EndIf
    \State Return ${T^1}$
\end{algorithmic}
\end{algorithm}

\subsubsection{Tree Building Algorithm} Taking Equation~(\ref{equation:kernel_distance_estimator}) as our split criterion, we aim to reduce the distance as small as possible. In this paper, we use $T^1$ (with a superscript 1) to indicate the causal balance tree (as well as branches) built in the first stage. As shown in Algorithm~\ref{code:causalbalance}, at an arbitrary parent node $T^1$ we split on the optimal partition $s_{T^1}^{\ast}$ that minimize:
\begin{equation}
    s_{T^1}^{\ast} = \mathrm{argmin}_{s_{T^1}} \frac{1}{N_T}\bigg( N_l \cdot \widehat{D}_{\mathcal{H}_1}[p_l,q_l] + N_r \cdot \widehat{D}_{\mathcal{H}_1}[p_r,q_r]\bigg),
    \label{equation:bestsplit}
\end{equation}
where $N_T, N_l, N_r$ indicate the sizes of parent, left child $l$, as well as right child $r$. Besides, $\widehat{D}_{\mathcal{H}_1}[p_l,q_l]$ and $\widehat{D}_{\mathcal{H}_1}[p_r,q_r]$ are kernel distance estimates for both children. When the chosen feature $k$ to split on is continuous, the candidate cutoff points are empirically determined by the distinct values of this feature; If $k$ is discrete, then the form of $Z_{k} \in A$ ($A$ is a subset of categories) induces the split. It is not possible to split the tree indefinitely, and stopping criteria based on tree complexity and node size are employed. Similar to existing decision tree algorithms \cite{laber2015tree}, we set a maximum depth of the tree $d_1$, the minimum number of units from either of the two treatment groups $n^1_{\mathrm{min}}, n^0_{\mathrm{min}}$, as well as the minimum leaf size $n_{\mathrm{min}}$. 
\yrv{Lines 3-7 in Algorithm~\ref{code:causalbalance} implement the {\it greedy recursive} tree-splitting procedure, iteratively partitioning the data to form subgroups while minimizing covariate imbalance, continuing until the stopping criteria are met. 
\xtengR{Stage 1 uses a greedy search strategy: it optimizes imbalance reduction locally at each split rather than evaluating all possible tree structures. A globally optimal tree, defined as the structure that minimizes imbalance under the same constraints (binary splits, recursive partitioning, finite depth), is computationally infeasible to compute. Greedy search is therefore the admissible approach for scalability. While it does not guarantee the global optimum, it consistently achieves large imbalance reductions in practice. We further elaborate our design choice in Section~\ref{sec:deparadox:deparadox_tree:time} as we discussed the time complexity of the algorithm.}}
% \xtengR{It is important to note that since this first stage follows a greedy search approach, it optimizes locally at each step but does not guarantee a globally optimal tree. A globally optimal tree is defined as the best structure, under the same tree constraints (e.g., binary splits, recursive partitioning, finite depth, finite number of leaves, etc), that yields the minimum total balance for a dataset. The recursive greedy strategy is computationally tractable but may diverge from the global optimum, since local imbalance reduction at early nodes do not always guarantee the best overall structure. In other words, greedy search minimizes imbalance {\it locally} through a sequence of nodes, while global search evaluates all tree configurations in the entire {\it entire} space and chooses the one that yields the minimum balance. In Section~\ref{sec:deparadox:deparadox_tree:time}, we will clarify our design that prioritizes scalability and interpretability, accepting local suboptimality in exchange for practical applicability. Besides, we discussed the time complexity of this first-stage algorithm, along with that of the second-stage algorithm, covering both stages.}}

\subsection{Stage 2: Growing An Opposite-Effects Policy Tree}\label{sec:deparadoxtree:policytree}
\begin{figure}[h]
    \centering
    \includegraphics[width=0.3\linewidth]{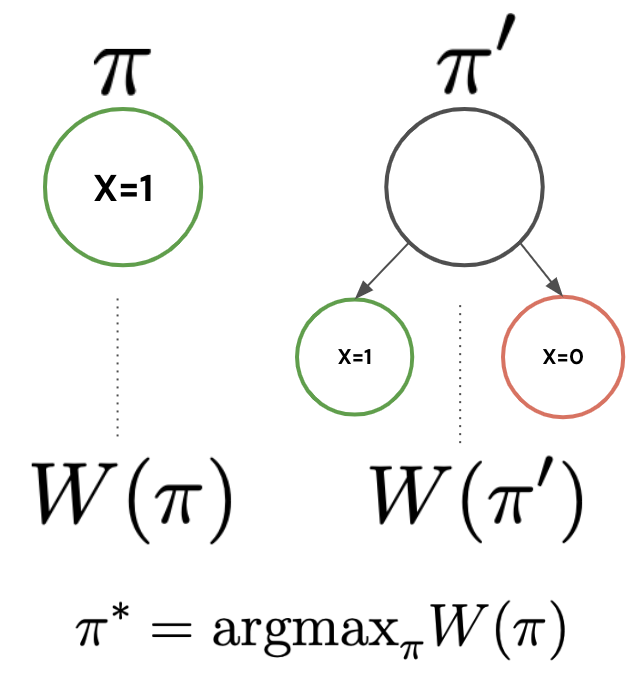}
    \caption[An illustration of policy learning to maximize expected outcomes.]{An illustration of policy learning to maximize expected outcomes. Suppose $\pi, \pi^{\prime}$ are two different policy trees, if the average outcome $W(\pi^{\prime}) > W(\pi)$, the second tree will be chosen and a root node will be further decomposed into two child subgroups with opposite treatments.}
    \label{fig:policy_tree}
\end{figure}

\revision{Suppose we have already obtained a set of balanced terminals; within each balanced leaf, we further expand an opposite-effects policy tree. In \deparadoxtree, we characterize a homogeneous subgroup to be a subgroup wherein all units should uniformly receive the same treatment for maximized expected outcome. In contrast, nested opposite effects occur when units within a group prefer opposite treatments to optimize the expected outcome. Figure~\ref{fig:policy_tree} depicts its core idea, where the left tree has a policy $\pi$ that assigns all samples into the treated $X=1$, whereas the right tree $\pi^{\prime}$ decomposes units into two subgroups with treatment $X=1$ versus $X=0$. The expected outcomes $W(\pi), W(\pi^{\prime})$ for two trees can be computed and compared to make a decision regarding tree selection.}

\subsubsection{Policy Tree} Following the principal of optimizing expected outcome, we adopt {\it policy learning} \cite{dudik2014doubly,zhou2023offline,kitagawa2018should,athey2021policy}, whose core idea is to learn an optimal {\it policy} $\pi^{\ast}$ that maps a covariate vector to a treatment action: $\pi^{\ast}: \mathcal{Z} \rightarrow \{0,1\}$, so that the expected outcome $W(\pi)$ could be maximized (illustrated in Figure~\ref{fig:policy_tree}):
\begin{equation}
    \pi^{\ast} = \mathrm{argmax}_{\pi}W(\pi) = \mathrm{argmax}_{\pi}\mathrm{E}_{\mathbf{Z}} \sum \big[\sum_{a\in\{0,1\}}Y(a, \mathbf{Z})\pi(a|\mathbf{Z}) \big],
    \label{equation:policylearning}
\end{equation}
\revision{where $a\in\{0,1\}$ is the potential treatments by following the target policy $\pi$ mapping features into two values---either 0 or 1. If $\pi(0|\mathbf{Z})=1$, it means the policy assign any unit with a feature vector $\mathbf{Z}$ to treatment $X=0$, and vice versa. A challenge in computing $W$ lies in that $Y(a,\mathbf{Z})$ is unknown to investigators particularly for the cases where the target treatment $a$ differs from the observed one $X_i$ ($a \neq X_i$). To overcome this obstacle, one needs to {\it estimate} the expected outcome of a given policy $\widehat{W}(\pi)$. Suppose $\widehat{e}(X|\mathbf{Z}) \triangleq p(X|\mathbf{Z})$ is an estimator of the conditional distribution of choosing a treatment based on covariates, referred as {\it propensity score model}; Besides, $\widehat{m}(X,\mathbf{Z})\triangleq p(Y|X,\mathbf{Z})$ is an estimator of the conditional mean response given treatment and feature values, referred as {\it response model}. Leveraging $\widehat{e}$ or $\widehat{m}$ (or both), there have been various methods proposed to estimate the expected outcome $\widehat{W}$ \cite{horvitz1952generalization,imbens2010rubin,beygelzimer2009offset,dudik2011doubly,dudik2014doubly}, such as direct method (DM) \cite{dudik2014doubly} that directly fits two response models $\widehat{m}(1,\mathbf{Z}), \widehat{m}(0,\mathbf{Z})$, and inverse property score (IPS) estimator \cite{horvitz1952generalization} which learns $\widehat{e}$ to re-weight each sample by the inverse probability of that sample receiving the treatment we observed $\frac{\pi(X|\mathbf{Z})}{\widehat{e}(X|\mathbf{Z})}Y$. In this paper, we use {\it doubly robust} (DR) estimator that contains both $\widehat{e}$ and $\widehat{m}$ \cite{dudik2014doubly}. It can be expressed as:
\begin{equation}
\begin{split}
    \widehat{W}^{\mathrm{DR}}(\pi) &= \frac{1}{N}\sum_i\widehat{W}^{\mathrm{DR}}_i(\pi),
\label{equation:drestimator}
\end{split}
\end{equation}
where each individual $\widehat{W}_i^{\mathrm{DR}}(\pi)$ can be written as:
\begin{equation}
\begin{split}
\widehat{W}_i^{\mathrm{DR}}(\pi)
&=\sum_{a\in\{0,1\}}\pi(a|\mathbf{Z}_i)\widehat{m}(a, \mathbf{Z}_{i}) + \frac{\pi(X_i|\mathbf{Z}_{i})}{\widehat{e}(X_i|\mathbf{Z}_{i})} \cdot \bigg(Y_{i} - \widehat{m}(X_i, \mathbf{Z}_i) \bigg) \\
&=\begin{cases}
 \text{$ \widehat{m}(0, \mathbf{Z}_{i}) + \frac{\delta(X_i=0)}{\widehat{e}(X_i|\mathbf{Z}_{i})} \cdot (Y_{i} - \widehat{m}(X_i, \mathbf{Z}_i) ) $}, & \text{if $\pi(0|\mathbf{Z}_i)=1$} \\
 \text{$ \widehat{m}(1, \mathbf{Z}_{i}) + \frac{\delta(X_1=1)}{\widehat{e}(X_i|\mathbf{Z}_{i})} \cdot (Y_{i} - \widehat{m}(X_i, \mathbf{Z}_i) ) $}, & \text{if $\pi(1|\mathbf{Z}_i)=1$}
 \end{cases}
\label{equation:drestimator_individual}
\end{split}
\end{equation}
It fits direct a regression model--$\widehat{m}(0,\mathbf{Z})$, or $\widehat{m}(1,\mathbf{Z})$ (the first term,), but then debiases that model by applying an inverse propensity approach to the residual of that model (the second term). This yields the overall algorithm: first we learn a response model $\widehat{m}$ by regressing outcome $Y$ on $X, \mathbf{Z}$, and a propensity score model by running a classification to predict $X$ from $\mathbf{Z}$. Then we construct the doubly robust estimator as above. The DR estimator holds advantages than DM or IPS methods because it is able to provide a {\it consistent and unbiased} estimates of $W(\pi)$ even when either $\widehat{e}$ or $\widehat{m}$ is misspecified \cite{dudik2014doubly}. Appendix~\ref{appendix:welfare_estimators} provides an example explaining this good property, along with more information about comparisons of the three estimators.}

Policies can manifest in various forms, such as a linear or nonlinear function or a set of rules. In our study, we specifically constrain the policy to a decision tree of a certain depth. By adhering to decision rules based on covariate values, individuals are categorized into specific leaves (subgroups). Subgroups recommended for treatment $X=1$ are expected to yield a positive effect, while those for treatment $X=0$ hold a negative effect. Notably, when two leaves have differing treatment assignments, they inherently represent opposing effects.

\begin{algorithm}[h]
\caption{\yrv{
$\mathrm{\bf OppositeEffectsPolicyTree}$: Identify nested subgroups with varying treatment preferences.
}}
\label{code:policytree}
\begin{algorithmic}[1]
  \Require \revision{Observational data $\mathcal{D}^{v}$ associated with an arbitrary balanced node $v$ from Algorithm~\ref{code:causalbalance}, the maximum depth of subtrees in the second stage $d^2$, and pre-computed $\widehat{m},\widehat{e}$, globally sort data samples along $K$ features and cache the indices}
  \If{$d=1$}
  \State Create a root node $T^2$ consisting of the entire $\mathcal{D}^{v}$
  \State Obtain the maximized empirical outcome $\widehat{W}^{\ast}$ and the best treatment action $\pi^{\ast}$ based on Equation~(\ref{equation:drestimator}) 
  \State Children nodes are empty $T^2_l \longleftarrow \emptyset, T^2_r \longleftarrow \emptyset$.
  \State Return $\big(\widehat{W}^{\ast}, {T^2} \big)$.
  \Else
  \State $\widehat{W}^{\ast} = -\infty, {T^2} \longleftarrow \emptyset$.
  \For{$k$ from 1 to $K$}
    \State \revision{\# Remember that data units have been sorted globally according to the $k$-th variable ${Z}_k$}.
    \For{$i$ from 1 to $N^v$}
    \State \revision{\# Recursively call the function itself to search for the best subtrees and outcomes}
    \State Let $\mathcal{D}^v_l=\{1,..., i\}$ and $\mathcal{D}^v_r=\{i+1,...,N^v\}$
    \State \revision{Obtain $(\widehat{W}^{\ast}_l,T^2_l) \longleftarrow \mathrm{OppositeEffectsPolicyTree}(\mathcal{D}^v_l)$ with depth $d^2-1$}
    \State \revision{Obtain $(\widehat{W}^{\ast}_r,T^2_r) \longleftarrow \mathrm{OppositeEffectsPolicyTree}(\mathcal{D}^v_r)$ with depth $d^2-1$}
    \If{$\widehat{W}^{\ast}_l + \widehat{W}^{\ast}_r > \widehat{W}^{\ast}$}
    \State $\widehat{W}^{\ast} \longleftarrow \widehat{W}^{\ast}_l + \widehat{W}^{\ast}_r$, {the best split} $s^{\ast} \longleftarrow (k, \frac{Z_{(i,k)}+Z_{(i+1,k)}}{2})$, ${T}^2 \longleftarrow \{s^{\ast}, ({T}^2_l, {T}^2_r)\}$.
    \EndIf
    \EndFor
  \EndFor
  \State Return $\big(\widehat{W}^{\ast}, {T}^2 \big)$.
  \EndIf
\end{algorithmic}
\label{algo:policytree}
\end{algorithm}

\revision{\subsubsection{Tree Building Algorithm} We use $T^2$ (with a superscript 2) to indicate the tree or branches built in the second stage. For each balanced node $v$ as well as its associated observable data $\mathcal{D}^v$ obtained from Algorithm~\ref{code:causalbalance}, the goal is to grow an {\it optimal} opposite-effects subtree $T^2$ and its {\it optimal} policy $\pi^{\ast}$ that maximizes Equation~(\ref{equation:drestimator}).}

\revision{As a preliminary step prior to tree building, we estimate two models $\widehat{m},\widehat{e}$ through a $C$-fold cross-fitted approach \cite{zhou2023offline,chernozhukov2018double}. Cross-fitting is a commonly used technique in statistical estimation to reduce model overfit and guarantee generalization \cite{chernozhukov2018double}. In our study, data is divided into $C$ folds, and the pair of $\{\widehat{m}^{-c}, \widehat{e}^{-c}\}$ is obtained through the $C-1$ folds of data except for the $c$-fold. Therefore, $\widehat{W}^{\mathrm{DR}}_i$ for data unit $i$ is computed based on $\{\widehat{m}^{-c},\widehat{e}^{-c}\}$ except for its own $c$-fold. Afterwards, Algorithm~\ref{algo:policytree} describes the procedure to search for an optimal opposite-effects policy tree of a certain depth $d$. In lines 1-5, if $d=1$ there no need to perform splitting, we can simply compute expected outcome $\widehat{W}^{\ast}$ and return the root node. Otherwise the problem decouples into two independent but smaller subproblems (lines 12-14): one for the left optimal subtree and the other for the right optimal subtree (a dynamic programming style algorithm) \cite{zhou2023offline}.} Different from the first stage of greedy search process, the second stage is able to obtain the optimal solution given tree depth $d$.

\subsection{The Proposed \deparadoxtree Algorithm}\label{sec:deparadox:deparadox_tree}
\begin{algorithm}[h]
\caption{
\yrv{Overall process of the \deparadoxtree algorithm.}
}
\label{code:deparadoxtree}
\begin{algorithmic}[1]
    \Require Observational data $\mathcal{D}$, the stopping criteria for Algorithm~\ref{code:causalbalance}, and Algorithm~\ref{code:policytree}.
    \State Run Algorithm~\ref{code:causalbalance} to obtain a tree ${T}^{1}$ consisting of a list of leaf nodes indexed by $v$
    \For{each leaf $v$}
    \State Run Algorithm~\ref{code:policytree} on $\mathcal{D}_v$ to grow ${T}_v^2$
    \EndFor
    \State Compute two node properties based on Equation~(\ref{equation:kernel_distance_estimator}) and Equation~(\ref{equation:effectsizeestimate}) for all nodes
    \State Return $({T}^1, \{{T}^2_v\})$
\end{algorithmic}
\end{algorithm}
\revision{The \deparadoxtree is a two-step procedure of growing a causal balance tree first (Algorithm~\ref{code:causalbalance}), and then growing opposite-effects policy subtrees from the balanced leaves (Algorithm~\ref{algo:policytree}). As shown in Algorithm~\ref{code:deparadoxtree}, line 1 corresponds to the first stage of tree building, and lines 2-4 to the second stage. Line 5 shows that, after a tree structure is obtained, we compute two node properties for every node $j$: a kernel distance $\widehat{D}_j$ following Equation~(\ref{equation:kernel_distance_estimator}), and a cause-outcome effect coefficient $\widehat{\rho}_{j}$ by fitting a linear regression function:
\begin{equation}
    Y = \rho\cdot X + u(\mathbf{Z}) + \epsilon,
\label{equation:effectsizeestimate}
\end{equation}
where $u(\mathbf{Z})$ is a linear term controlling the residual imbalance, and $\epsilon$ is assumed to be Gaussian noises.}

\subsubsection{Time Complexity Analyses}\label{sec:deparadox:deparadox_tree:time} In Algorithm~\ref{code:deparadoxtree}, line 1 is a greedy tree search procedure, whereas line 3 is an optimal tree search algorithm. They have distinct runtime complexities. 
\begin{enumerate}
    \item {\bf Algorithm~\ref{code:causalbalance}---CausalBalanceTree.} The requirement line shows that, to find the best split we need to sort $N$ data samples based on $K$ features. The sorting has time complexity $N\mathrm{log}N$, if we have to compare $K$ features, this becomes $O(KN\mathrm{log}N)$. Line 4 for computing kernel distance has a runtime cost $O(n_1 + n_0)^2$ where $n_0 + n_1$ is the sample size of a node. However, in our implementation we precomputed a $\kappa$ matrix that scales to $O(N^2)$ which reduces line 4 to $O(N)$. In the case where $T^1$ will not be skewed binary tree, its runtime cost would be $O({N}^2 + K{N}\mathrm{log}{N})$.
    \item {\bf Algorithm~\ref{algo:policytree}---OppositeEffectsPolicyTree.} In the second stage, line 8-10 also require that data samples are globally sorted along $K$ dimensions. Fortunately we can inherit the globally sorted indices could be shared by both stages thus resorting could be avoided. The recursive step in line 8-19 is a dynamic programming process which scales with $O({N^v}^{d^2}K^{d^2}\mathrm{log}{N^v})$ \cite{sverdrup2020policytree} where $d^2$ is subtree depth in stage 2 and $N^v$ is sample size of nodes obtained from stage 1.
\end{enumerate}
Aggregating two stages, it shows that the proposed \deparadoxtree can be solved in polynomial time. \xtengRp{The polynomial runtime of the greedy search reflects its role as the tractable solution to an otherwise NP-hard problem \cite{laurent1976constructing}. Here, the dominant cost arises from the one-time construction of the ($N \times N$) kernel matrix, while subsequent greedy splitting involves comparatively low overhead due to shallow tree depth.} Global optimization would require searching exponentially many tree configurations, which is infeasible for realistic datasets. {Greedy search, by contrast, enables application to moderate-scale observational datasets while still yielding meaningful imbalance reduction. \xtengRp{On a single machine, datasets on the order of several thousand to tens of thousands of samples are feasible without approximation. For larger datasets, scalability can be extended through standard approximations such as subsampling or low-rank kernel methods, which we leave as future work.}

\subsubsection{Further Notes Applying \deparadoxtree}
\yrv{
Applying the proposed method to real-world datasets requires domain knowledge to ensure accurate causal interpretation. The method is best suited for scenarios where Simpson’s paradox arises due to imbalance or heterogeneity (as depicted in Figure~\ref{fig:causal_structure}). It is not applicable in cases where Simpson’s paradox results from colliders or more complex causal structures beyond those considered in this study \cite{hernan2011simpson}.

\paragraph{When to apply the method?}
\begin{enumerate}
    \item {\bf Datasets with potential confounding due to covariate imbalance.} If treatment and control groups differ in key covariates, Simpson’s paradox may indicate a spurious association, and our method can help identify meaningful subgroups where the effect is more reliably estimated.
    \item {\bf Datasets with potential heterogeneous treatment effects.} When the overall effect masks subgroup-specific variations, our method can uncover groups with opposing or varying effects, improving policy or decision-making.
    \item {\bf Exploratory analysis in causal inference.} Even when Simpson’s paradox is not observed, the method can verify whether the overall association remains robust across subgroups or whether balancing covariates alters the estimated effect.
    \item {\bf Well-balanced covariates \& homogeneous effects.} If a dataset exhibits well-balanced covariates or a homogeneous treatment effect, no paradoxical trend reversal will be detected. In such cases, the root node itself serves as the final \deparadoxtree, meaning the overall association is a reliable indicator of the causal effect. However, if Simpson’s paradox is suspected despite observed balance, further robustness checks should be performed to ensure that unobserved confounding is not influencing the results.
\end{enumerate}

\paragraph{When the method may not be necessary?} 
\begin{itemize}
    \item {\bf Causal structures involving collider bias.} Our method is not suitable for datasets where Simpson’s paradox arises due to collider bias, which occurs when a variable is conditioned upon that is influenced by both the exposure and the outcome, leading to distorted associations \cite{hernan2011simpson}. To assess the potential presence of collider bias before applying our method, researchers should consider causal graphical analysis using Directed Acyclic Graphs (DAGs) to identify colliders and avoid conditioning on them \cite{pearl2000simpson, pearl1995causal,greenland1999causal}. Additionally, sensitivity analyses, such as E-values and simulation-based approaches, can help estimate the potential impact of unmeasured colliders on causal estimates \cite{vanderweele2017sensitivity}. Additionally, Structural Equation Modeling (SEM) can be used to explicitly model direct and indirect pathways, allowing researchers to assess whether a collider is distorting the observed effect \cite{kline2023principles}. If collider bias is suspected, alternative causal modeling techniques should be used to ensure valid inference.
\end{itemize}

\paragraph{How to interpret the results?}
\begin{enumerate}
    \item \xtengR{{\bf If a square-node causal balance tree is produced:}}
The method successfully partitions the data into balanced subgroups, indicating that the association reversal was driven by imbalance rather than a genuine causal effect.
{\bf Interpretation:} The observed effect in the balanced subgroups can be considered more reliable than the overall association.
    \item \xtengR{{\bf If a circular-node opposite-effects policy tree is produced without a causal balance tree:}}
The method reveals that treatment effects vary across subgroups, even when covariates are balanced.
{\bf Interpretation:} Simpson’s paradox in this case does not necessarily indicate spuriousness but rather suggests that the treatment effect is heterogeneous across different groups.
    \item \xtengR{{\bf No paradoxical trend reversal is detected, and the \deparadoxtree output a single root node:}}
If no trend reversal is detected after applying the method, the dataset may already exhibit well-balanced covariates, meaning that Simpson’s paradox is not present due to imbalance or heterogeneity. In such cases, the root node itself serves as the final \deparadoxtree, indicating that the overall association is likely a valid representation of the treatment effect. However, the absence of a trend reversal does not guarantee the absence of unmeasured confounders or collider bias, which are not directly addressed by this method.
{\bf Interpretation:} If no subgroup effect variations are detected, this suggests that Simpson’s paradox was not a concern under the assumed causal structure. However, if theoretical expectations or prior research suggest that a trend reversal should occur, the absence of a detected reversal may indicate the influence of unmeasured confounders or collider distortions rather than true effect homogeneity.
{\bf Further checks} include:
(a) Assess unmeasured confounding: Even if observed covariates are balanced, hidden confounders could still bias the results. Sensitivity analyses, such as E-values or quantitative bias analysis, can help estimate the potential impact of these confounders on causal conclusions.
(b) Check for collider bias: If a paradoxical association was expected but not observed, it may be due to conditioning on a collider variable that distorts causal relationships. Conducting a Directed Acyclic Graph (DAG) analysis can help identify whether a collider effect might be present and misleading the results.
(c) Verify causal structure assumptions: If results contradict prior knowledge, revisiting the assumed causal structure and considering alternative DAG specifications can help determine if omitted pathways or unintended variable conditioning are affecting the findings.
\end{enumerate}
Following these guidelines, data practitioners can apply \deparadoxtree more effectively and ensure that the detected associations are interpreted correctly within their causal context.
}

\section{Experiments and Results}\label{sec:results}
This section introduces baseline decision trees in causal inference literature. \xtengR{We evaluate our method through a simulation experiment, a hybrid experiment, and analyses of two real-world applications.}

\subsection{Baseline Methods}
\yrv{
We selected five state-of-the-art decision tree models from the fields of {\bf causal inference} and {\bf uplift modeling} based on their effectiveness in estimating treatment effects and subgroup heterogeneity. These models were chosen for their ability to either (1) {\bf effectively partition data} or (2) {\bf preserve treatment effect heterogeneity}, making them suitable baselines for evaluating the robust estimation and interpretation of the causal effects of our approach. These include:
% \xtengR{We excluded methods in Section~\ref{sec:simpson_paradox} \cite{alipourfard2021dogr,alipourfard2018can, alipourfard2018using,xu2018detecting,wang2023learning}  as they are fundamentally different from the proposed \deparadoxtree (such as, single-variable disaggregation \cite{alipourfard2018can, alipourfard2018using} vs multi-variable, soft clustering \cite{alipourfard2021dogr} vs hard disaggregation, statistical trends \cite{alipourfard2021dogr,alipourfard2018can, alipourfard2018using,xu2018detecting,wang2023learning} vs causal effects}). 
\begin{enumerate}
    \item {\bf Causal Inference Trees}:
    \begin{itemize}
        \item {\bf Propensity Tree} \cite{wager2018estimation} leverages estimated propensity scores to construct treatment effect heterogeneity-aware trees. It recursively splits data and puts units with similar propensity scores into the same subgroups, due to the theoretical proof that homogeneous subsets regarding propensity score yield the same covariate distributions for treated and control groups \cite{rosenbaum1983central}. It aims to mitigate confounding bias and thus enhance accurate causal effect estimation in subgroups. 
        \item {\bf Causal Tree} \cite{athey2016recursive}: A recursive partitioning approach designed to minimize variance in treatment effect estimation. It was designed to uncover heterogeneity in treatment effects without relying on a pre-defined analysis plan. It recursively divides data into subgroups to minimize the expectation of the mean squared error (MSE) with respect to a causal effect estimator. It divides tree-building and treatment effect estimation into two steps. Samples are divided into distinct subsets for placing splits and estimating causal effects, respectively. This two-step procedure helps eliminate bias in treatment effect estimation.
        \item {\bf Interaction Tree} \cite{su2008interaction, su2009subgroup} identifies subgroups with significant treatment effect modifications based on interaction terms. It is also designed to explore the heterogeneity structure of the treatment effect across a number of subgroups that are objectively defined in a post-hoc manner \cite{su2008interaction,su2009subgroup}. At each step, it selects the best split that shows the greatest interaction between split and treatment. The measure for assessing the covariate-treatment interaction is proven to be equivalent to the $t$ test for testing a null hypothesis that no interaction effect exists in data.
        \item {\bf CATE Tree Interpreter} \cite{econml} constructs interpretable decision trees for the conditional average treatment effect (CATE), aiming at offering a clearer understanding of effect heterogeneity. It provides a presentation-ready summary of the key features that explain the differences in the responses to an intervention in the presence of confounding variables \cite{econml}. Before a tree could be built, conditional causal effects need to be estimated through any of the available CATE estimators. Then the model splits on the cutoff points that maximize the treatment effect difference in each leaf. Our experiments test on two popular estimators: double machine learning (DML) \cite{chernozhukov2018double} and doubly robust learners (DRL) \cite{foster2019orthogonal}, as both have leveraged machine learning techniques to model both the outcome and the treatment assignment to better account for confounding bias. Depending on which ML models being used, we selected {\it CausalForest-DML}, {\it Linear-DML} and {\it Linear-DRLearner}\footnote{\url{https://econml.azurewebsites.net/spec/spec.html}} from the above two categories \cite{econml}.
    \end{itemize}
    \item {\bf Uplift Modeling Trees}:
    \begin{itemize}
        \item {\bf Uplift Trees} \cite{gutierrez2017causal, radcliffe2011real} optimize splits based on the gain in expected treatment outcome responses. They are also known as uplift modeling or treatment effects trees, which are a type of predictive modeling techniques used in marketing and personalized intervention strategies \cite{rzepakowski2012decision,gutierrez2017causal,radcliffe2011real,chen2020causalml}. The goal of uplift modeling is to identify the individuals or segments of a population that are most likely to respond positively to a treatment or intervention. Our experiments considered two types of split criteria: information theory-based, e.g., {\it Uplift Tree-KL}, {\it Uplift Tree-ED}, and {\it Uplift Tree-Chi} \cite{radcliffe2011real}, and Contextual Treatment Selection {\it Uplift Tree-CTS} \cite{zhao2017uplift}. The former measures divergence between outcome distributions, while the latter directly maximizes the gain in expected response. We conducted experiments using the Causalml package\footnote{\url{https://github.com/uber/causalml}} \cite{chen2020causalml}.
    \end{itemize}
\end{enumerate}
}

\xtengR{We did not include non-decision-tree baselines, as they are not directly comparable to our setting. Existing methods \cite{alipourfard2018can, alipourfard2018using,xu2018detecting,wang2023learning,alipourfard2021dogr} differ fundamentally in scope and assumptions: \cite{alipourfard2018can, alipourfard2018using,xu2018detecting} only detect Simpson's paradox caused by a single variable, unlike our multi-variable framework; \cite{alipourfard2021dogr} search for overlapping clusters, while our approach focuses on non-overlapping disaggregation; \cite{alipourfard2018can, alipourfard2018using,xu2018detecting,wang2023learning,alipourfard2021dogr} identify trend reversals but ignores causal mechanisms behind the reversal.}
\xtengR{For illustration, we applied method \cite{alipourfard2018using} to the hybrid dataset (see Section~\ref{sec:results::hybrid}, Appendix ~\ref{appendix:nontree}). While it successfully identified one Simpson's pair, it did not capture the injected imbalance and heterogeneous effects across four features. This highlights the gap between simple pairwise reversal detection methods and the broader causal disaggregation addressed by our method.}

\subsection{Experiments on Synthetic Datasets}\label{sec:results::synthetic}
\begin{figure}[h]
    \centering
    \includegraphics[width=0.45\linewidth]{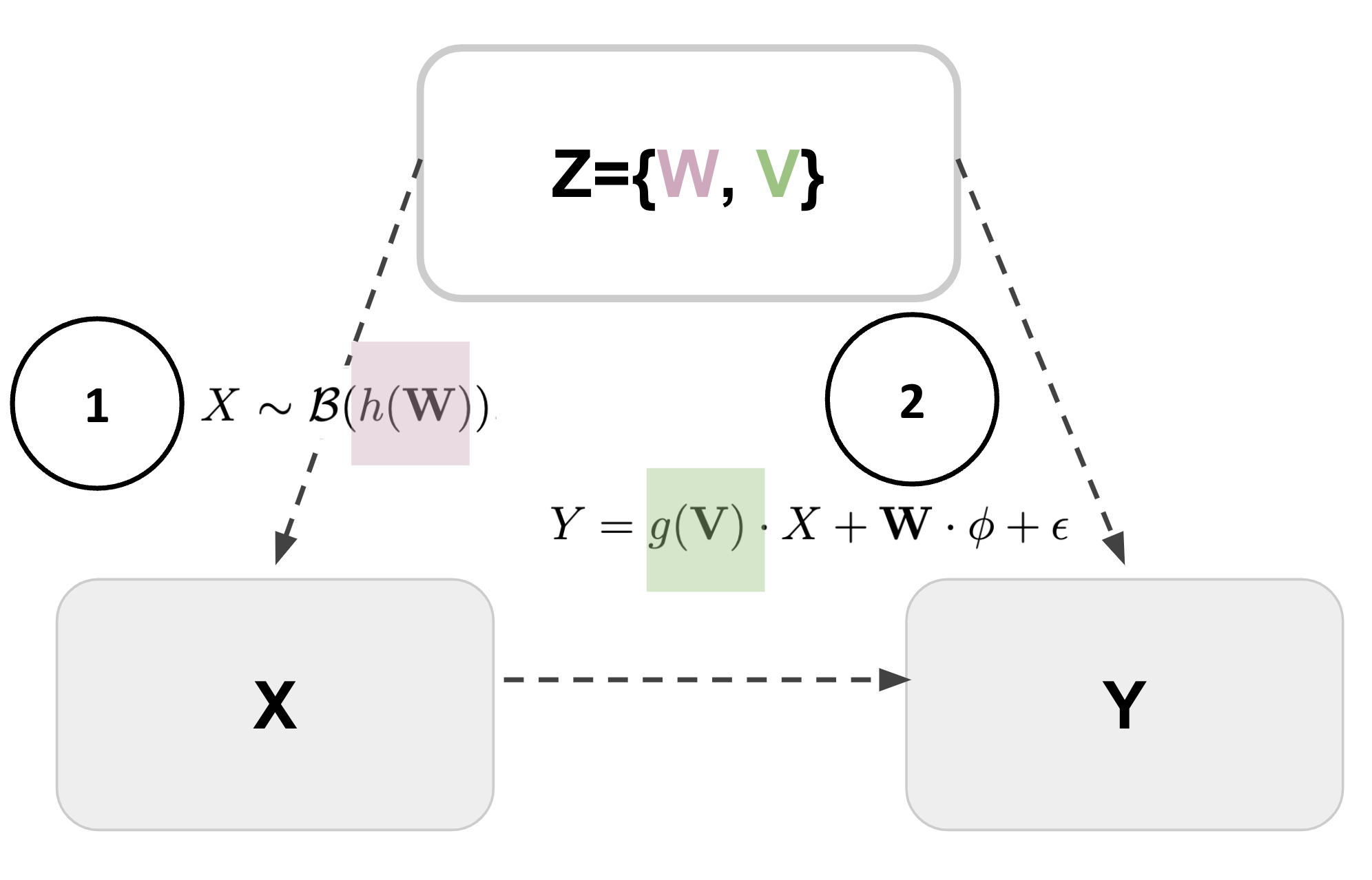}
    \caption[A two-step data simulation process to inject imbalance and heterogeneity.]{The two-step data simulation process, where $h(\mathbf{W})$, defined in Equation~(\ref{eq:propensityscore}) based on $f(\mathbf{W})$, determines the injected imbalance, as well as $g(\mathbf{V})$, seen in Equation~(\ref{eq:outcomeregression}), determines the injected heterogeneity. (1) We design a function $f(\mathbf{W})$ corresponding to {\bf $f$-Design 0}, {\bf $f$-Design 1}, and {\bf $f$-Design 2} to create three variants of imbalance; (2) We design a function $g(\mathbf{W})$ corresponding to {\bf $g$-Design 0}, {\bf $g$-Design 1}, and {\bf $g$-Design 2} to create three variants of heterogeneity.}
    \label{fig:data_simulatioin}
\end{figure}

\subsubsection{Data Simulation}
We conducted a simulation study incorporating various designs of imbalance and effect heterogeneity. We utilized a hypothetical data generation process, assuming covariates $\mathbf{Z}$ could contain confounders $\mathbf{W}$ that controls imbalance and features $\mathbf{V}$ that determines heterogeneity. Data simulation process can be seen in Figure~\ref{fig:data_simulatioin}. In step (1), the binary treatment variable $X$ follows a Bernoulli distribution, $X \sim \mathcal{B}(h(\mathbf{W}))$, the origin of treatment imbalance. In other words, $X$ is a random variable which takes the value 1 with probability $h(\mathbf{W})$ and the value 0 with probability $1 - h(\mathbf{W})$. The probability function $h(\mathbf{W})$ is a logistic function of $\mathbf{W}$,
\begin{equation}
    h(\mathbf{W}) = \frac{1}{1+e^{-f(\mathbf{W})\alpha}},
    \label{eq:propensityscore}
\end{equation}
where $\alpha$ is the parameter. As $\mathbf{W}$ determines to what extent an arbitrary unit would or would not get the intervention, function $f$ (and $\alpha$) determines what the imbalance patterns would be like over two treatment groups. We considered three imbalance designs, corresponding to three functions $f(\mathbf{W})$:
\begin{enumerate}
    \item {\bf $f$-Design 0}: $f(\mathbf{W}) = 1$ suggesting that there is no confounders at all. That is, the Bernoulli distribution has a constant treatment probability $h$ independent of $\mathbf{W}$.
    \item {\bf $f$-Design 1}: $f(\mathbf{W}) = \mathbbm{1}(\mathbf{W} \ge 0) - \mathbbm{1}(\mathbf{W} < 0)$, suggesting that subgroups with $\mathbf{W} \ge 0$ and $\mathbf{W} < 0$ would hold two distinct values of Bernoulli probability, i.e., $1/(1+e^{-\alpha})$ versus $1/(1+e^\alpha)$.
    \item {\bf $f$-Design 2}: $f(\mathbf{W}) = \mathbf{W}$, suggesting a nonlinear relationship between $\mathbf{W}$ and treatment $X$.
\end{enumerate}
In step (2), outcome $Y$ is assumed to be produced following an additive function:
\begin{equation}
    Y = g(\mathbf{V})\cdot X + \mathbf{W}\cdot\phi + \epsilon,
    \label{eq:outcomeregression}
\end{equation}
where $\epsilon \sim \mathcal{N}(0,1)$ and $g(\mathbf{V})$ captures the effect heterogeneity patterns across subgroups defined by features $\mathbf{V}$. Similarly, we considered three designs of heterogeneous effects: 
\begin{enumerate}
    \item {\bf $g$-Design 0}: $g(\mathbf{V}) = 1$, suggesting a homogeneous effect.
    \item {\bf $g$-Design 1}: $g(\mathbf{V}) = \Big[\mathbbm{1}(\mathbf{V} \ge 0) - \mathbbm{1}(\mathbf{V} < 0)\Big] \cdot \beta$. Subgroups with $\mathbf{V} \ge 0$ and $\mathbf{V} < 0$ would hold two different values of effect sizes: $\beta$ versus $-\beta$. There is a clear boundary at $\mathbf{V} = 0$.
    \item {\bf $g$-Design 2}: $g(\mathbf{V}) = \mathbf{V}\cdot \beta$. It suggests that the effect magnitude is linearly dependent on $\mathbf{V}$.
\end{enumerate}
Together, Equation~(\ref{eq:propensityscore}) and Equation~(\ref{eq:outcomeregression}) are the true propensity score model and true response model. For each of the designs, we randomly drew parameter values from Gaussian distribution $\alpha \sim \mathcal{N}(0,1), \beta \sim \mathcal{N}(0,1)$, and $\phi \sim \mathcal{N}(0,1)$. We generated 100 datasets based on 100 sets of randomly-sampled parameter values $\{\alpha_i, \beta_i, \phi_i\}$. The total size of units is set to be 2000.

\subsubsection{Performance Metrics}
\yrv{
We use multiple metrics to assess different aspects of the solutions, which can be categorized into {\bf interpretability}, {\bf balance}, {\bf causal inference accuracy}, and {\bf policy evaluation}.
\begin{enumerate}
    \item {\bf Interpretability}: Tree size, measured by the {\bf number of leaves}, represents the number of subgroups in a tree. A {\bf smaller} number is preferred, as smaller trees are easier to interpret and generalize better.
    \item {\bf Covariate Balance}: {\bf Kernel balance}, measured by kernel distance, and the {\bf Kolmogorov-Smirnov (KS) statistic}, assess how well the subgroups in the tree are balanced. A {\bf smaller} value for these metrics indicates that covariates $\mathbf{Z}$ are better adjusted between treated and control arms, reducing confounding bias.
    \item {\bf Causal Inference Accuracy}: 
    \begin{itemize}
        \item {\bf Mean squared error (MSE)} quantifies the accuracy of treatment effect estimation by measuring the average squared difference between the estimated and true causal effect. {\bf Lower} MSE indicates higher estimation accuracy.
        \item {\bf Coverage} evaluates uncertainty by measuring the proportion of cases where the 95\% confidence interval of the estimated effect includes the true causal effect. A {\bf higher} coverage value indicates a more reliable estimation method.
    \end{itemize}     
    \item {\bf Policy Evaluation for Heterogeneous Effects}: 
    \begin{itemize}
        \item {\bf Regret} quantifies how suboptimal a learned policy tree $\pi$ is compared to the optimal policy $\pi^{\ast}$, computed as the difference in cumulative reward: $W(\pi^{\ast}) - W({\pi})$. In data simulations, $W({\pi^{\ast}})$ is known, while $W({\pi})$ is computed for each learned tree. {\bf Lower} regret indicates a more effective policy.
        \item {\bf Error rate} measures the fraction of incorrectly predicted treatments compared to ground-truth treatments. A {\bf smaller} error rate signifies better treatment assignment accuracy.
    \end{itemize}
\end{enumerate}
}

\subsubsection{Results}

\begin{figure}[h]
    \centering
    \includegraphics[width=\linewidth]{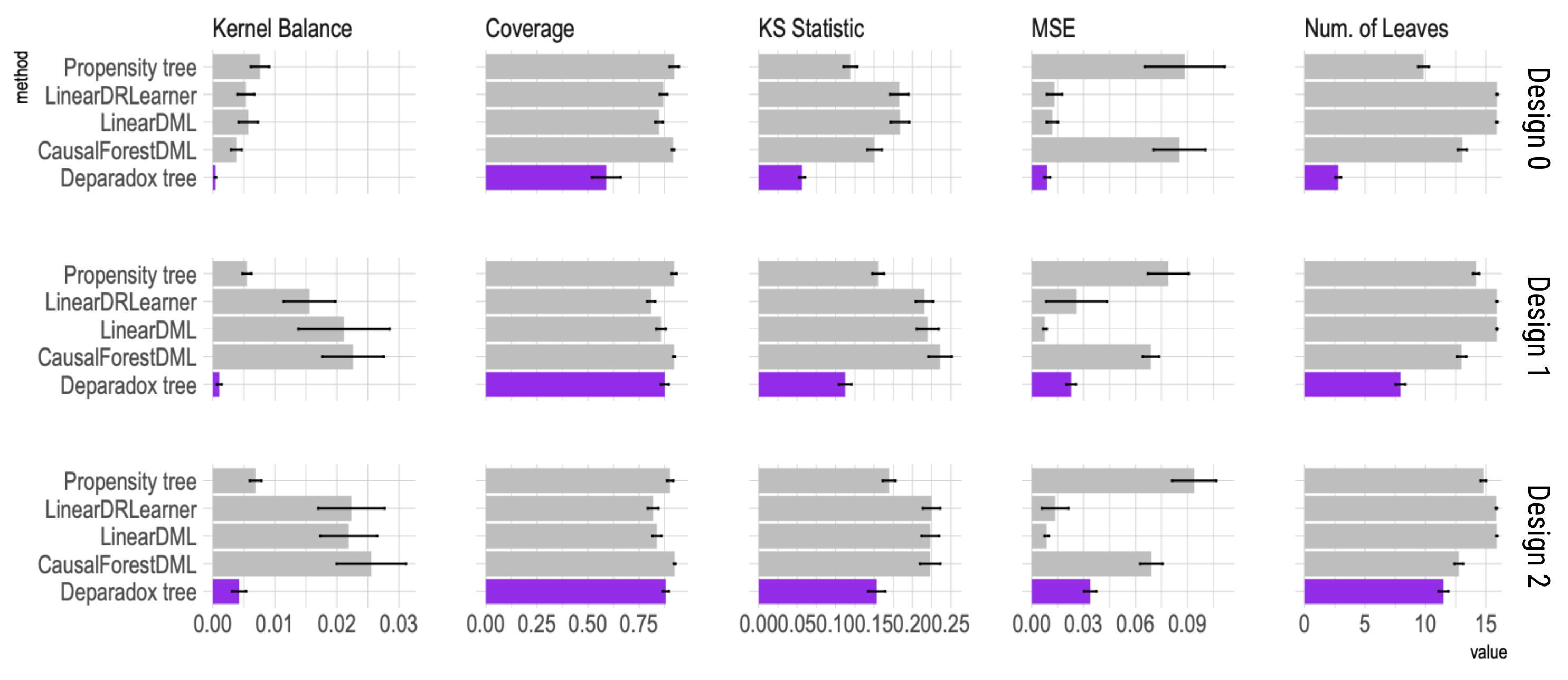}
    \caption[Comparing \deparadoxtree and baseline methods on simulation data with injected confounder imbalance.]{Comparing \deparadoxtree and baseline methods on simulation data with injected confounder imbalance---{\bf $f$-Design 0, $f$-Design 1, $f$-Design 2}. Three rows represent three designs of confounder imbalance. We measure how balanced the subgroups are through subgroup kernel balance, and average Kolmogorov-Smirnov (KS) test, how accurate the subgroup-level effect estimates are through coverage rate and MSE of causal effects, and how many subgroups generated using the number of leaves. In each design we report the average and 95\% CI from 100 simulation trials with $N=2000$ data units. Tree configuration parameters are $n_{\mathrm{min}} = 30, n^1_{\mathrm{min}} = n^0_{\mathrm{min}} = 15, d_1 = 4, d_2 = 4$.
    }
    \label{fig:compr_f}
\end{figure}
\begin{figure}[h]
    \centering
    \includegraphics[width=0.7\linewidth]{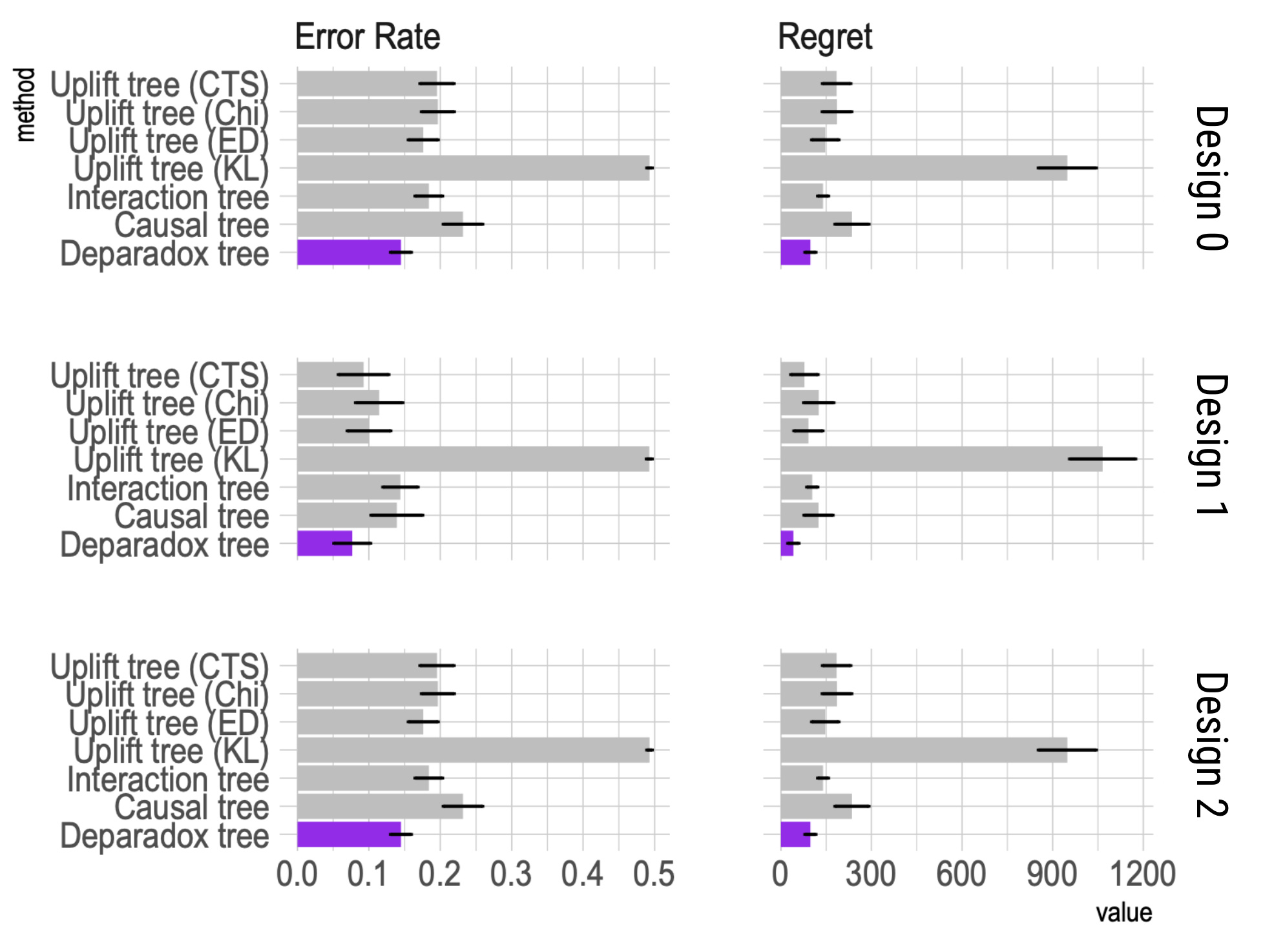}
    \caption[Comparing \deparadoxtree and baseline methods on simulation data with injected heterogeneous effects.]{Comparing \deparadoxtree and baseline methods on simulation data with injected heterogeneous effects. Three rows suggest three designs of heterogeneous effects---{\bf $g$-Design 0, $g$-Design 1, $g$-Design 2}. We measure how accurate the assigned treatments are through error rate of treatment assignments and the regret between optimal outcome and the outcome achieved by tested methods. The smaller values the better. The dimensions of $\mathbf{W}, \mathbf{V}$ are both 2. In each design we report the average and 95\% CI from 100 simulation trials with $N=2000$ data units. Tree configuration parameters are $n_{\mathrm{min}} = 30, n^1_{\mathrm{min}} = n^0_{\mathrm{min}} = 15, d_1 = 4, d_2 = 4$.
    }
    \label{fig:compr_g}
\end{figure}

\xteng{Figure~\ref{fig:compr_f} reports the results---both mean and 95\% CI from 100 simulations---for three imbalance designs $f$ with $N=2000$ and no effect heterogeneity inserted (i.e., $g(\mathbf{V})=1$).} The proposed algorithm yields shallower trees than baseline methods. Moving from {\bf $f$-Design 0} to {\bf $f$-Design 2}, the average number of leaves increases, {indicating a larger number of subgroups are discovered to achieve homogeneous effects. That is, more complicated imbalance pattern results in a deeper and more complex tree structure.} Investigating the columns of two balance metrics---kernel balance and KS statistics---reveals that the proposed algorithm achieves lower imbalance scores than baseline methods, aligning with the proposed split criterion's balancing objective. The margin between the proposed method and the propensity tree is significant, indicating that modeling the propensity score through decision trees might not ensure a well-balanced covariate distribution in practice. According to the MSE and coverage columns in Figure~\ref{fig:compr_f}, the proposed algorithm outperforms the propensity tree and even two CATE tree---{\it CausalForest-DML} and {\it Linear-DRLearner}---specifically designed for causal effect estimation in the presence of confounders. {\it LinearDML} outperforms the proposed method as it holds the lowest MSE, however its imbalance scores (see kernel balance and KS statistic) are high, suggesting that the subgroup-level associations are distorted and cannot be used for making decisions. Figure~\ref{fig:compr_g} displays results---both mean and 95\% CI from 100 simulations---for three designs of effect heterogeneity $g$ with $N=2000$ and no confounders (i.e., $f(\mathbf{W}) = 1$). The results demonstrate that the proposed method exhibits the lowest error rate and regret among all methods, benefiting from its well-designed objective of maximizing global welfare and its split criterion of searching for the global optimal tree policy.

\subsection{Experiments on A Hybrid Voting Dataset}\label{sec:results::hybrid}
\subsubsection{Data Simulation} 

We further assessed the proposed method's performance in scenarios with both imbalance and heterogeneity using a hybrid voting dataset. The voting dataset, originally collected by Gerber et al. \cite{gerber2008social}, and includes 180,002 data points representing individual voters across 344,082 households in Michigan. Eight voter characteristics, including {\it year of birth}, {\it sex}, {\it household size}, {\it g2000}, {\it g2002}, {\it p2000}, {\it p2002}, and {\it p2004}, serve as features. These features are averaged within households, therefore, {\it sex} indicates the percentage of female family members, {\it year of birth} implies average age, and {\it g2000}, {\it g2002}, {\it p2000}, {\it p2002}, {\it p2004} shows the percentage of family members who have voted for general and primary elections in 2000, 2002, and 2004 respectively. The target outcome is whether a person voted in the August 2006 primary election. In the original study, households were randomly assigned to either the control group or the Civic Duty treatment group, where a letter encouraging civic duty (i.e., ``Do your civic duty'') was emailed before the primary election \cite{gerber2008social}. 

% Our experiments focus on the Civic Duty treatment, i.e., a letter with ``Do your civic duty'' is emailed to the household before the primary election, comparing it to the Control condition. 

To create a hybrid dataset, we intentionally introduced both imbalance and effect heterogeneity into the data, following a similar simulation approach as in Zhou et al. \cite{zhou2023offline}. We aimed to evaluate whether the proposed method can accurately identify the injected patterns. In this simulation, we envisioned that the Civic Duty letters were assigned with varying probabilities to different subgroups based on two features: {\it year of birth} and {\it household size}. We defined three subgroups as follows: 
\begin{enumerate}
    \item People in households with an average age \underline{younger} than the median and a household size \underline{smaller} than the median would have a high chance of being mailed the Civic Duty letter.
    \item People in households with an average age \underline{younger} than the median and a household size \underline{larger} than the median would have a high chance of being assigned to the control condition.
    \item People in households with an average age \underline{older} than the median are also likely assigned to the control condition.
\end{enumerate}
We processed each data point by transforming its original treatment status. If the original treatment matches the likely treatment described above, we kept it; Otherwise, we flipped the original treatment with a probability $p_{X_{\mathrm{flip}}}$.

To inject effect heterogeneity, we further assumed different subgroups of people respond differently to the Civic Duty letter and specified 3 subgroups defined by another two features: {\it p2004} and {\it sex}.
\begin{enumerate}
    \item People in households with a voting rate \underline{lower} than the median in the 2004 primary election and a \underline{lower} percentage of female family members than the median prefer the Civic Duty letter over the control. In other words, these individuals are more likely to vote when receiving a Civic Duty letter.
    \item People in households with a voting rate \underline{lower} than the median in the 2004 primary election and a \underline{higher} percentage of female family members than the median prefer the control over the Civic Duty letter. Namely, these individuals are less likely to vote when receiving a Civic Duty letter.
    \item People in households with a voting rate \underline{higher} than the median prefer the control over the Civic Duty letter.
\end{enumerate}
We processed each data point by transforming its original outcome status to a virtual one. If the treatment matches the preferred one and the original outcome is $Y=0$, we flipped the it from $Y=0$ to $Y=1$ with a probability $p_{Y_{\mathrm{flip}}}$; If the treatment does not match the preferred one and the original outcome is $Y=1$, we flipped it from $Y=1$ to $Y=0$ with a probability $p_{Y_{\mathrm{flip}}}$. 

\subsubsection{Performance Metrics}

\revision{When comparing \deparadoxtree against baseline method, we particularly looked at whether the produced trees split on the correct variables to reveal the injected data patterns. In balanced tree generation, we look forward to seeing that {\it year of birth} and {\it household size} are picked up as split variables; whereas in homogeneous tree splitting, we look forward to seeing {\it p2004} and {\it sex} as the two selected features.}

\subsubsection{Results}

\begin{figure}[h]
    \centering
    \includegraphics[width=\linewidth]{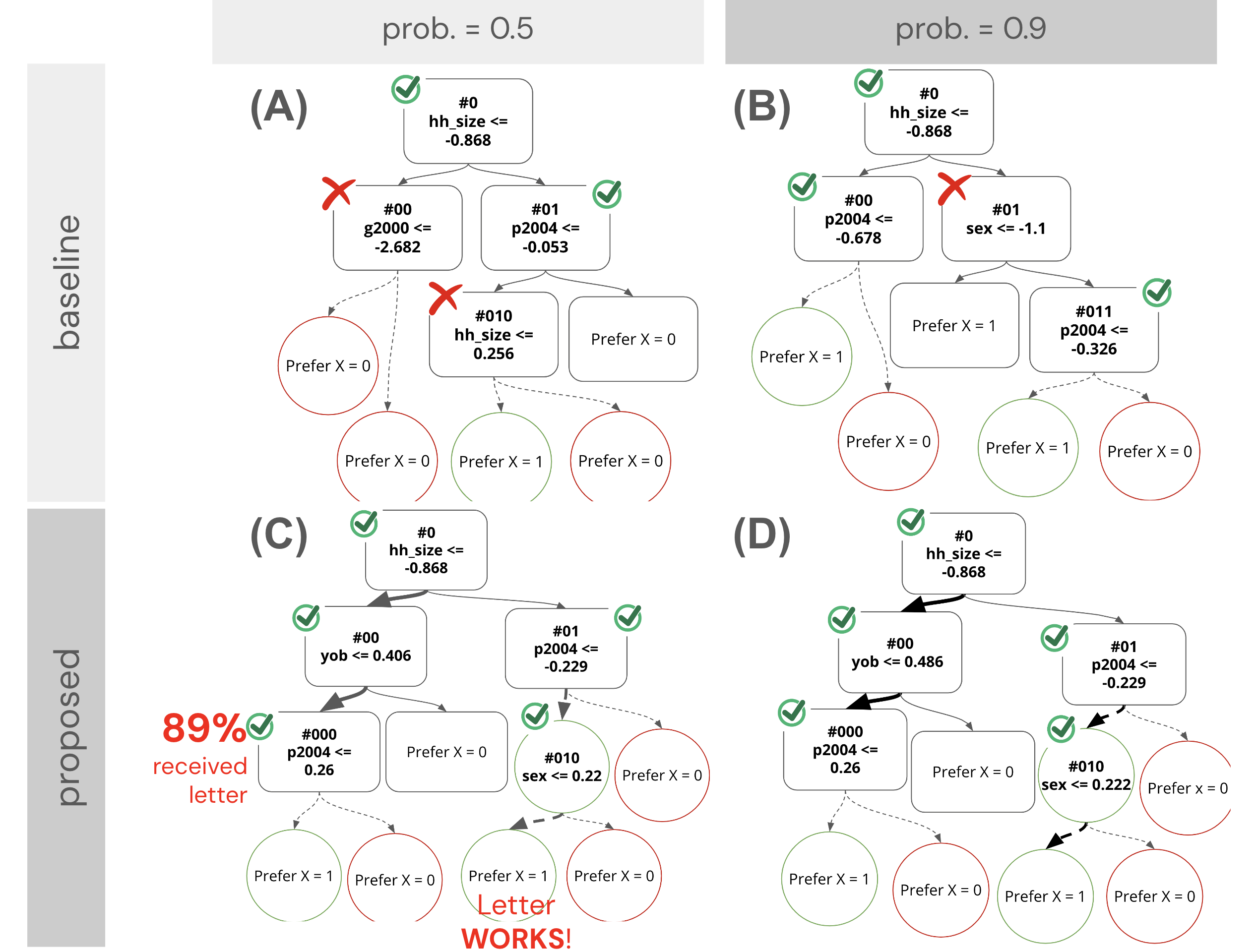}
    \caption[Comparing \deparadoxtree vs baseline method on the hybrid voting datasets.]{Comparing trees from baseline (A, B) on hybrid voting data and our \deparadoxtree (C, D). For (A, C), $p_{{X_\mathrm{flip}}} = 0.5$ and $p_{{Y_\mathrm{flip}}} = 0.5$, while for (B, D), $p_{{X_\mathrm{flip}}} = 0.9$ and $p_{{Y_\mathrm{flip}}} = 0.9$. Node boxes include ID, split criterion, kernel balance, effect size with 95\% CI, and sample size for treated (NT) and control (NC) groups. Square nodes are to mitigate confounding bias, circular nodes are to search mixed effects. Marked nodes match the ground-truth variable for subgroup specification. Same tree configurations are used for both \deparadoxtree and baseline $n_{\mathrm{min}} = 30, n^1_{\mathrm{min}} = n^0_{\mathrm{min}} = 15, d_1 = 4, d_2 = 2$.
\yrv{
{\bf Explanation:} The comparison shows that \deparadoxtree outperforms baseline methods by correctly identifying key splitting variables to reduce imbalance and reveal heterogeneous treatment effects. It selects {\it household size} and {\it year of birth} for balancing, then uncovers {\bf nested subgroups} using ground-truth features. In contrast, baselines often split on incorrect variables, missing key confounders and effect modifiers. Even with stronger treatment imbalance (flip probability 0.9), baselines still overlook critical features, demonstrating the robustness of \deparadoxtree in complex observational data.
}
    }
    \label{fig:hybridcompr}
\end{figure}
We varied the flip probabilities and generate three versions of hybrid voting datasets: both $p_{X_{\mathrm{flip}}}=p_{Y_{\mathrm{flip}}}=10\%$, $p_{X_{\mathrm{flip}}}=p_{Y_{\mathrm{flip}}}=50\%$, and $p_{X_{\mathrm{flip}}}=p_{Y_{\mathrm{flip}}}=90\%$, corresponding to different strengths of treatment imbalance and effect heterogeneity: weak, medium, strong. Figure~\ref{fig:hybridcompr} compares trees from baseline (A, B) vs our \deparadoxtree (C, D) on hybrid voting data. Subplots (A, C) indicate the medium case where flipping probabilities are 0.5, subplots (B, D) show the strong case where where flipping probabilities are 0.9.\footnote{Refer to Figure~\ref{fig:appendix_hybridcompr} at Appendix~\ref{appendix:appendix_hybridcompr} for the weak case where flipping probabilities are 0.1.} The baseline method consists a propensity tree and a uplift tree (ED).\footnote{{We chose propensity tree as it is particularly designed for adjusting confounders in observational data while the other methods are originally proposed for randomized trials. We selected the ED-based uplift tree because it gives consistent small values regarding regret and error rate in Figure~\ref{fig:compr_g}.}} Figure~\ref{fig:hybridcompr} (C, D) shows that our proposed method has successfully located {\it household size} (\texttt{hh\_size}) at Node \texttt{\#0} and {\it year of birth} (\texttt{yob}) at Node \texttt{\#00} as two split features to reduce imbalance. This procedure successfully finds a subset of households---small and young---where 89\% of which have received the Civic Duty letter. After the first stage, the proposed \deparadoxtree further reveals the nested subgroups that hold opposite treatment effects by splitting on the ground-truth splitting features {\it p2004} at Node \texttt{\#01} and {\it sex} at Node \texttt{\#010}. This procedure successfully finds a subset of households, with a lower-than-median voting rate and less female members, are more likely to vote after receiving a Civic Duty letter. 
In contrast, the trees produced by baseline methods split on incorrect features. In Figure~\ref{fig:hybridcompr} (A), it misses {\it year of birth} as the splitting variable to reduce imbalance, instead it wrongly selects {\it g2000} at Node \texttt{\#00}. Besides, it fails to identify the variable {\it sex} we have used to inject differential effects, instead it selects {\it hh\_size} at Node \texttt{\#010}. Unfortunately, the split criterion at Node \texttt{\#00} produces two children nodes that have the same treatment preference. In Figure~\ref{fig:hybridcompr} (D), as flipping probabilities increase to 0.9, the baseline is able to locates more correct splitting variables, but it still overlooks \texttt{yob} for balancing and {\it sex} for homogenizing. {A detailed copy of Figure~\ref{fig:hybridcompr} can be seen in Figure~\ref{fig:hybridcompr_original} at Appendix~\ref{appendix:appendix_hybridcompr}.}

% \xtengR{We also ran the hybrid data on a non-tree baseline ~\cite{}.\footnote{Source code: https://github.com/ninoch/Trend-Simpsons-Paradox}}

\subsection{Application on the Real-World Datasets}\label{sec:results::realistic}
\xtengR{We analyzed two real-world datasets from different domains. Through \deparadoxtree, a confounder-imbalance-driven paradox was detected in the Lalonde dataset (Section ~\ref{experiments:realworld:lalonde}), and a heterogeneity-driven paradox was revealed in the Python programming dataset (Section ~\ref{experiments:realworld:python}). Together, these analyses demonstrate the flexibility of our approach.}

\subsubsection{The Lalonde Dataset}\label{experiments:realworld:lalonde}
\begin{figure}[h]
    \centering
    \includegraphics[width=\linewidth]{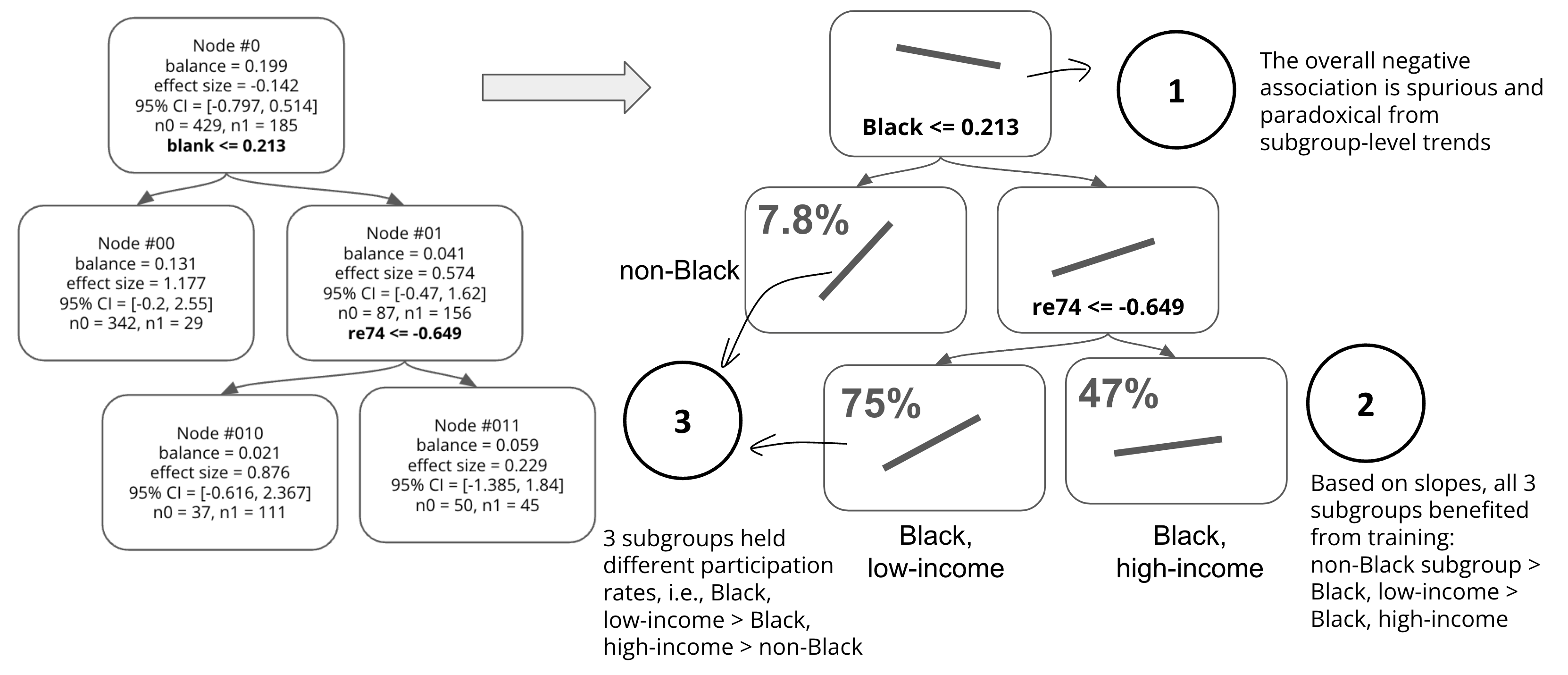}
    \caption[Applying the proposed \deparadoxtree algorithm on the Lalonde dataset.]{The tree learned from the \deparadoxtree algorithm when applied on the Lalonde dataset. The left tree shows detailed statistics, the right one is for paradox investigation. Slopes represent effect size, percentages are participation rates. Tree parameters are $d_1 = 4, d_2 = 2, n_{\mathrm{min}} = 30, n^1_{\mathrm{min}}=n^0_{\mathrm{min}}=15$.
\yrv{
{\bf Explanation:} The results show that \deparadoxtree effectively uncovers subgroup-level treatment effects in this dataset, revealing meaningful heterogeneity that traditional analyses overlook. While the overall effect appears spurious (-0.142, 95\% CI: [-0.797, 0.514]), subgroup analysis {\bf identifies positive treatment effects across all subgroups}, with the non-Black subgroup benefiting the most. Additionally, kernel balance improves after partitioning, reinforcing the reliability of subgroup-level estimates. These findings highlight the method’s interpretability and potential for uncovering hidden disparities in treatment response.
}
}
    \label{fig:lalonde_deparadox}
\end{figure}

We applied the proposed \deparadoxtree to the Lalonde dataset, a benchmark in causal inference literature \cite{lalonde1986evaluating,dehejia2002propensity,dehejia1999causal}. This dataset was obtained from the federally-funded National Supported Work (NSW) Demonstration in the mid-1970s, aimed at providing working experience to individuals who were facing economic and social difficulties pre-enrollmnet. It includes pretreatment variables such as pre-treatment earnings in 1974 and 1975 ({\tt re74}, {\tt re75}), {\tt age}, education years ({\tt educ}), ethnicity ({\tt black}, {\tt hispanic}), marital status ({\tt married}), and educational degree ({\tt nodegree}). The outcome variable is post-training earnings in 1978 ({\tt re78}). We utilized a subset of the dataset, as curated by \cite{lalonde1986evaluating,dehejia1999causal,dehejia2002propensity}, focusing on male participants assigned between Jan 1976 and July 1977 with earnings data in 1974. This subset comprises 185 treated and 260 control observations. To create a non-randomized dataset, we introduced a comparison dataset, called Westat's Matched Current Population Survey-Social Security Administration File (CPS-3). The treatment group consists of participants in the training program, while the control group is CPS data (CPS-3) \cite{lalonde1986evaluating,dehejia1999causal,dehejia2002propensity}. Overall, we have 185 treated and 429 control units.\footnote{Feature values are transformed thus different from its raw values.} Figure~\ref{fig:lalonde_deparadox} illustrates subgroups identified by the proposed algorithm using two split features, {\tt black} and {\tt re74}, yielding three subgroups: (1) Non-Black (Node \texttt{\#00}, 371 persons), (2) Black, low-income (Node \texttt{\#010}, 148 persons), and (3) Black, high-income (Node \texttt{\#011}, 95 persons). The participation rates for these subgroups are 7.8\% (29 out of 371), 75\% (111 out of 148), and 47\% (45 out of 95), respectively, indicating a lower interest among non-Black individuals and a higher interest among Black, low-income individuals.

By investigating the generated \deparadoxtree, three RQs {\bf 1--3} can be answered accordingly. The overall negative effect size for the entire Lalonde dataset is -0.142 (95\% CI: [-0.797, 0.514]), deemed spurious due to confounding (answering {\bf RQ 1}). The tree splits on two confouding variables---{\tt black} and {\tt re74}. When controlled for, reversed positive trends emerge at the three causal balanced subgroups (square nodes)--1.177 (95\% CI: [-0.2, 2.553]) for Node \texttt{\#00}, 0.875 (95\% CI: [-0.616, 2.367]) for Node \texttt{\#010}, and 0.229 (95\% CI: [-1.385, 1.843]) for Node \texttt{\#011}. By following the tree structure, we observe paradoxical associations. The Black and non-Black square nodes exhibit reversed trends relative to their parent node. The Black, low-income and the Black, high-income subgroups exhibit different levels of positive trends. There are no further circular nodes generated in Figure~\ref{fig:lalonde_deparadox}, suggesting that there are no effect modifiers causing opposite effects within these three balanced subgroups (answering {\bf RQ 2}). After data partition, kernel balance decreases from the initial 0.199 to 0.131 for Node \texttt{\#00}, 0.021 for Node \texttt{\#010}, and 0.059 for Node \texttt{\#011}. This reduction in kernel balance from parent nodes to children nodes suggests that subgroup-level effect sizes are a more reliable estimations of true effects. When it comes to decision-making, the \deparadoxtree suggests that subgroup-level trends are more reliable than the overall as imbalance has been mitigated, therefore the job-training program seems to be beneficial for all three subgroups in improving people's incomes (answering {\bf RQ 3}). Overall, the tree is simple and interpretable: it has the potential to help user grasp the critical hidden information regarding different propensity levels and effect sizes across subgroups, e.g., the low-income Black node is the most engaged subgroup probably due to their desire to overcome financial obstacles, and non-Black node seems to benefit the most from the job training program with significant fairness implications.

\subsubsection{A Python Programming Dataset}\label{experiments:realworld:python}
\begin{figure}[h]
    \centering
    \includegraphics[width=\linewidth]{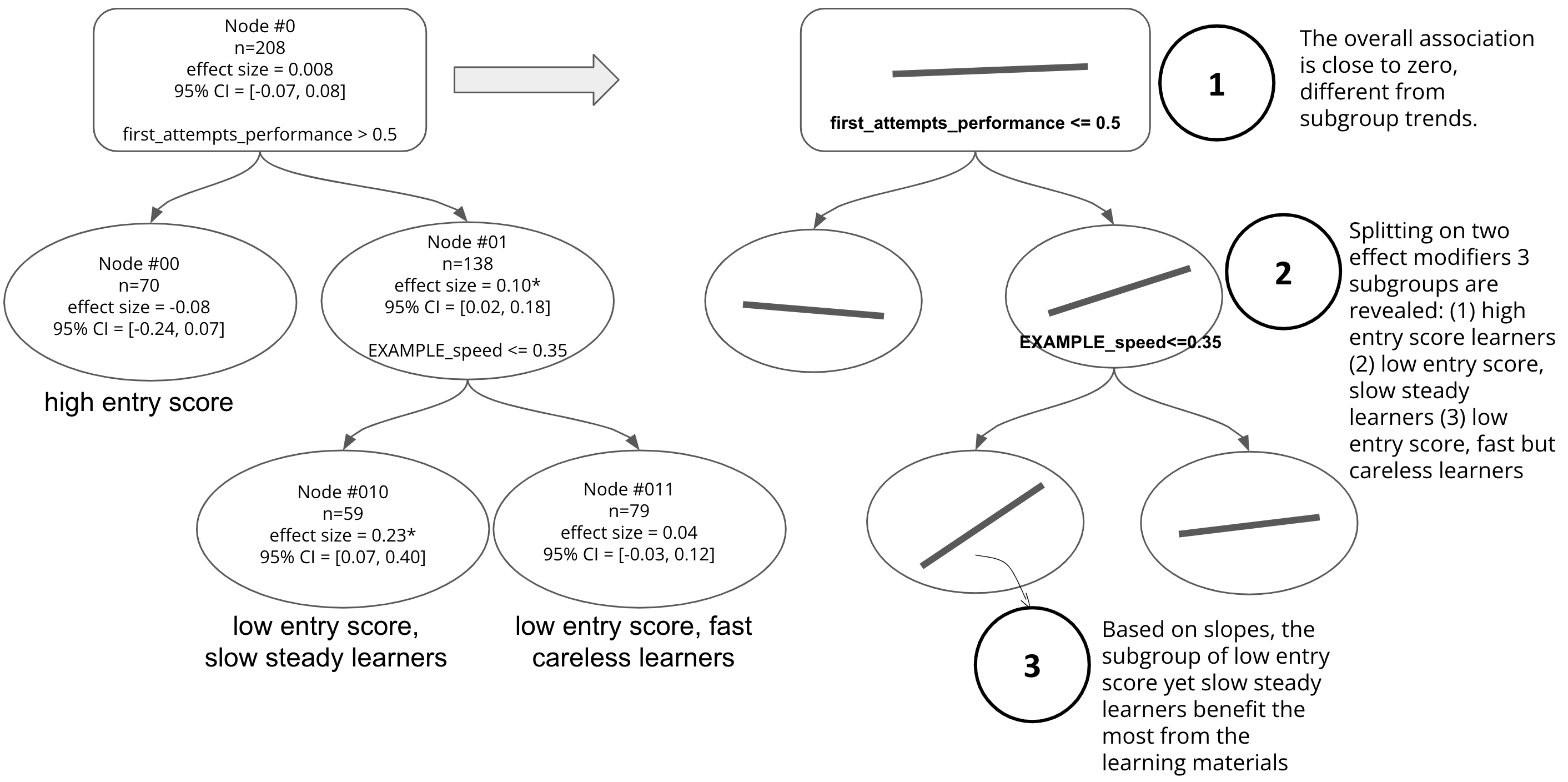}
    \caption[Applying the proposed \deparadoxtree algorithm on the Python Programming dataset.]{\xtengR{The tree learned from the \deparadoxtree algorithm when applied on the Python Programming dataset. The left tree shows detailed statistics, the right one is for paradox investigation. Slopes represent effect size, circle nodes mean it's an opposite-effects policy tree by splitting on two effect modifiers ${\tt first\_attempts\_performance}, {\tt EXAMPLE\_speed}$ Tree parameters are $d_1 = 4, d_2 = 2, n_{\mathrm{min}} = 30, n^1_{\mathrm{min}}=n^0_{\mathrm{min}}=15$. {\bf Explanation:} The results show that \deparadoxtree effectively uncovers subgroup-level treatment effects in this dataset, revealing meaningful heterogeneity that traditional analyses overlook. While the overall effect is close to zero (0.008, 95\% CI: [-0.07, 0.08]), subgroup analysis {\bf identifies positive treatment effects for low-entry-score students}, with the low-entry-score yet slow steady learners benefiting the most (0.23, 95\% CI: [0.07, 0.40]).}}
    \label{fig:python_deparadox}
\end{figure}
\xtengR{We also applied \deparadoxtree on an internal dataset collected from a Python programming classroom. This dataset comes from an online learning tool that provides code examples to help students understand Python programming concepts and structures. The tool was offered as an optional learning resource, and students used it voluntarily to explore examples and complete quiz tasks at their own pace, according to their needs and schedules. The dataset contains longitudinal interaction traces from 482 anonymized students across ten classes, recorded between February 2020 and October 2021. The raw data includes student ID, timestamp, task type (e.g., quiz or code example), task topic, steps taken to complete a task, correctness of the response, and time spent. In total, the dataset comprises 113,669 logged student actions, representing approximately 666 hours of tool usage. From this interaction data, we derived several static variables: {\tt EXAMPLE\_total\_attempts} – total number of attempts at code examples, {\tt EXAMPLE\_total\_duration} – total time spent on code examples, {\tt EXAMPLE\_speed} – average speed of completing a code example, {\tt first\_attempts\_performance} – number of correct answers in the first three quiz tasks, {\tt QUIZ\_performance} – overall quiz accuracy, {\tt QUIZ\_total\_attempts} – total attempts on quiz tasks, {\tt QUIZ\_speed} – average speed of completing a quiz.}

\xtengR{The main analysis examined whether system engagement—measured by the total number of code examples studied ({\tt EXAMPLE\_total\_number})—was associated with student performance, measured by quiz accuracy ({\tt QUIZ\_performance}). Additional variables such as {\tt first\_attempts\_performance} capture prior knowledge (entry score), while {\tt EXAMPLE\_speed} reflects the average time learners spent per material item. Figure \ref{fig:python_deparadox} shows a near-zero effect at the root node, suggesting that the overall number of examples studied was not significantly associated with performance—contrary to the developers’ expectations. However, the first split on {\tt first\_attempts\_performance} revealed two distinct subgroups:
\begin{enumerate}
\item {\bf High-entry-score learners} (Node \#00, 70 students), with a negligible effect size of -0.08 (95\% CI: [-0.24, 0.07]);
\item {\bf Low-entry-score learners} (Node \#01, 138 students), with a significant, positive effect size of 0.10 (95\% CI: [0.02, 0.18]).
\end{enumerate}
The positive and statistically significant effect for low-entry-score learners suggests that the system may have been particularly effective in helping students with weaker prior knowledge improve their skills. Within the low-entry-score subgroup, {\tt EXAMPLE\_speed} further distinguished two profiles:
\begin{enumerate}
\item {\bf Slow, steady learners} (Node \#010, 59 students), showing a strong positive effect size of 0.23 (95\% CI: [0.07, 0.40]);
\item {\bf Fast but careless learners} (Node \#011, 79 students), with a negligible effect size of 0.04 (95\% CI: [-0.03, 0.12]).
\end{enumerate}
Taken together, this analysis provides evidence that:
\begin{enumerate}
\item {\bf RQ1:} Overall effects are near zero, but heterogeneous subgroups exist;
\item {\bf RQ2:} Entry score and example speed act as effect modifiers;
\item {\bf RQ3:} The system was most effective for low-entry-score, slow steady learners, while no clear benefits were observed for high-entry-score or fast but careless learners.
\end{enumerate}}
\section{Conclusion and Discussion}

In conclusion, this study addresses the challenges posed by spurious associations in observational datasets and machine learning tools, particularly focusing on the phenomenon of Simpson's paradox. We introduced a novel data-driven method called the \deparadoxtree, which utilizes regression tree techniques to recursively partition data into subgroups, effectively mitigating the spuriousness of associations. The proposed approach not only builds simpler trees but also selects relevant covariates as splitting rules, creates balanced subgroups, and unveils nested opposite effects. Through comprehensive comparisons with state-of-the-art baseline methods on synthetic, hybrid, and real-world datasets, we demonstrated the effectiveness of the \deparadoxtree approach in achieving good coverage and small mean squared error of causal effects. Importantly, the proposed method is designed to be interpretable and user-friendly for both non-experts of causal inference and machine learning. By providing a clear tree summary of relevant subgroups and illustrating the gradual mitigation of imbalance and heterogeneity, the proposed approach empowers individuals to make more informed decisions based on subgroup-level associations.

\subsection{Tree as an interpretable method to investigate spurious associations and Simpson's paradox}
As discussed in the related work (ref. Section~\ref{sec:causal_subgroup_identification}), decision trees have been popular in causal inference and personalized intervention/marketing domains in empirical studies. By exploring the splits in the tree, researchers are able to uncover heterogeneity among interpretable subgroups, e.g., subgroups with different treatment effects, different preferences towards treatment, or different welfares/benefits considering treatment costs and responses. Despite of its popularity in observational data analysis, this paper is the first work, to our best knowledge, to employ decision tree for the purpose of studying spurious/paradoxical associations. \xteng{In light of this purpose, we design the partitioning criteria based on two causal structures depicted in Figure~\ref{fig:causal_structure}. The two mechanisms---imbalance and heterogeneity---are fundamental to understanding and resolving Simpson's paradox because they directly affect how aggregated associations differ from subgroup-specific causal effects. The de-paradox tree structure operationalizes these principles by: (1) Tracing trend reversals at different levels of partitioning; (2) Identifying key splitting variables that act as confounders (causing imbalance) or effect modifiers (causing heterogeneity); (3) Providing an interpretable structure that clearly illustrates when, where, and why Simpson’s paradox occurs.} \xtengRp{The tool is designed as a diagnostic and explanatory tool under assumed causal structures involving confounders and effect modifiers (Figure~\ref{fig:causal_structure}) whereas collider-based structures are intentionally out of scope. This is because balancing or splitting on collider variables implicitly conditions on them and can introduce selection bias, leading to spurious subgroup effects rather than valid causal explanations. Further work should consider improving the visual system (such as a warning box or disclaimers) to proactively communicate these causal requirements and limitations to users, encouraging them to reason about the role of candidate variables before including them in the split variable set.} The proposed method is also limited by disadvantages of decision trees. The structure of trees can be sensitive to the dataset: a slight change in the data may lead to different tree structures, potentially impacting the partition rules and estimated effects. \xtengRp{Because the tree is constructed via greedy search with axis-aligned splits, it may underfit when confounding arises from complex or highly nonlinear interactions among covariates, leaving residual imbalance that is not fully removed by partitioning alone. While the kernel distance used in our split criterion can capture higher-order and nonlinear discrepancies between covariate distributions, it cannot fully compensate for limited split expressiveness. This is an intentional design choice, as we prioritize interpretable causal decomposition over exhaustive bias minimization.} Furthermore, the method operates only on observable covariates, and they may not effectively handle unobservable confounders. If there are latent variables influencing both treatment assignment and outcomes, the trees may produce biased results.

% The tree performs automatic variable selection in discovering balanced subgroups and effect reversal, which is beneficial to enhance the awareness of selected confounders (and effect modifiers) and the distortions/paradox caused by those features.

\subsection{The relationship between two objectives: mitigating confounding bias and searching for effect reversal}
Our research concentrates on addressing confounding bias and subgroup heterogeneity, the primary drivers of Simpson's paradox. As our goal is to minimize the spuriousness of a subgroup-level association, we pursue two objectives: mitigating confounding bias, and searching for nested opposite effects. These two objective are difficult to be optimized simultaneously: when searching for homogeneous subgroups in terms of treatment effects, it is not guaranteed the covariates are balanced; Conversely, when searching for balanced subgroups, it is not guaranteed that no nested opposites effects are present. To address this challenge, our approach prioritizes mitigating confounding bias over identifying heterogeneous effects. Once balanced terminals are identified, the tree extends to explore potential effect reversal. This strategic choice underscores the reliance of the heterogeneous effects policy tree on the preceding causal balance tree.

\subsection{The trade-offs in \deparadoxtree}
{\it The trade-off between tree complexity and effect estimation.} In the realm of prediction-based machine learning, managing the complexity of decision trees is crucial for achieving a balance between overfitting training data and ensuring generalizability to test data. Common strategies include cross-validation and cost-complexity pruning \cite{quinlan1986induction}, where cost reflects prediction error and complexity relates to tree size. Analogously, there is also a trade-off between cost and complexity in the proposed \deparadoxtree method. The cost might manifest as residual confounders within a balanced terminal or nested effect reversal within a policy node. Our approach gives users the freedom to tailor tree size by specifying parameters like depth and minimal node size, offering flexibility for domain experts to align with their dataset knowledge. Moreover, through the specification of diverse parameters, users can conduct multiple experimental trials, allowing for a comparative analysis of tree structures and results to gain additional insights. It's crucial to note that the \deparadoxtree is not intended as a replacement for other causal inference trees and techniques. Rather, it serves as a low-cost, user-friendly method for non-experts to identify, investigate, and understand Simpson's paradox.

\xtengR{{\it The trade-off between scalability and global optimum.} Our method is built around a fundamental trade-off between scalability and global optimality. In principle, global optimization would identify the tree with minimum imbalance across all possible recursive partitions under the given constraints, but this becomes intractable for datasets beyond simple, easily separable cases. Greedy search is therefore the practical alternative: it ensures polynomial runtime and interpretability while still achieving substantial imbalance reduction.}

\xtengR{In practice, divergence from the global optimum usually occurs when an early local decision prevents more effective downstream splits. Three types of situations are most common. (1) Hidden interaction effects -- when two (or more) variables interact in a way that only becomes evident after both are split. A greedy procedure may bypass such a split, whereas a global search would capture the joint effect. (2) Minority–majority imbalance -- when covariate values vary disproportionately in small subgroups compared to a dominant majority. (3) Multi-step imbalance patterns -- certain variables only reduce imbalance after multiple splits (e.g., Age at 40 then 60). Greedy may instead choose a single intermediate split that provides short-term gains but blocks the more effective sequence.}

\xtengR{Despite these limitations, the method remains robust across diverse datasets. Future work could investigate approximation bounds or hybrid strategies to further narrow the gap with global optimization. One promising direction is to study data-driven, anytime approximation guarantees for greedy search. Similar to bounds in decision tree learning \cite{domingos2000mining}, node-wise lower bounds on achievable imbalance reduction could be compared with the actual cost of greedy splits. Such bounds would adapt to the data distribution--looser for easily separable subsets, tighter for harder-to-separate cases—and offer a more informative measure of solution quality.}

\section{Acknowledgments}
The authors would like to acknowledge support from AFOSR, ONR, Minerva, NSF \#2318461, as well as Pitt Cyber Institute's PCAG awards and Pitts' CRC resources. Any opinions, findings, and conclusions or recommendations expressed in this material do not necessarily reflect the views of the funding sources.

\bibliographystyle{ACM-Reference-Format}
\bibliography{references}

\appendix
\section{\deparadoxtree Methodologies}
\subsection{Kernel Methods}\label{appendix:kernels}
In machine learning, a kernel is a function that takes in two inputs (such as vectors) and calculates a measure of similarity between them. It tells us how ``close'' or ``similar'' two items are based on their features. Suppose two inputs can be described by two vectors $\mathbf{z}, \mathbf{z}^{\prime}\in\mathcal{Z}$, their inner product $\langle \mathbf{z}, \mathbf{z}^{\prime} \rangle$ is a similarity measure. However, the class of linear functions induced by this inner product may be too restrictive for many real-world problems. Kernel methods aim to build more flexible and powerful learning algorithms by transforming $\mathbf{z} \in \mathcal{Z}$ in the original space to a {\it feature space} $\phi(\mathbf{z})\in \mathcal{F}$. Hence, the natural extension of $\langle \mathbf{z}, \mathbf{z}^{\prime} \rangle$ is
\begin{equation}
    \kappa(\mathbf{z}, \mathbf{z}^{\prime}) = \langle \phi(\mathbf{z}), \phi(\mathbf{z}^{\prime}) \rangle_{\mathcal{F}},
\end{equation}
where $\langle\cdot,\cdot\rangle_{\mathcal{F}}$ denotes the inner product in the feature space $\mathcal{F}$. We refer to $\phi$ and $k$ as a {\it feature map} and a {\it kernel function}. 
% Consider an example of document classification, a kernel can tell us how similar two documents (collections of words) are based on their vectors of word frequency $\phi(\mathbf{z}), \phi(\mathbf{z}^{\prime})$ by mapping from a document space $\mathcal{Z}$ into a vector feature space $\mathcal{F}$. 
% However, the mapping step can be computationally expensive if $\phi(\mathbf{z})$ lives in a high-dimensional feature space. 
Consider a polynomial feature map $\phi(\mathbf{z}) = (x_1^2,x_2^2,\sqrt{2}x_1x_2)$ when $\mathbf{z}\in\mathbb{R}^2$ from \cite{gretton2013introduction}. Then we have
\begin{equation}
    \kappa(\mathbf{z}, \mathbf{z}^{\prime}) = z_1^2{z_1^{\prime}}^2 + z_2^2{z_2^{\prime}}^2 + 2z_1z_2{z_1^{\prime}}{z_2^{\prime}} = \langle\mathbf{z}_1,\mathbf{z}_2\rangle^2.
\end{equation}
The kernel is simply the square of the inner product of two vectors in the original space $\mathcal{Z}$. The equality holds for a more general $d$-degree polynomial and these two-step computation would be expensive when $d$ grows very large. An essential aspect of kernel methods is {\it kernel trick}: instead of performing obtaining $\phi(\mathbf{z})$ explicitly, it is sufficient to compute $\kappa(\mathbf{z}, \mathbf{z}^{\prime})=\langle\mathbf{z},\mathbf{z}^{\prime}\rangle^d$ directly in the original space.

\subsection{Reproducing Kernel Hilbert Space (RKHS)}\label{appendix:rkhs}
A {\it Hilbert space} is a space on which an inner product is defined. The elements in a Hilbert space could be vectors or {functions} \cite{muller1997integral}. For instance, let $\mathcal{H}$ be a {\it Hilbert space of functions} where the functions $f$ take values in the space of $\mathcal{Z}$, namely $f: \mathcal{Z} \longrightarrow \mathbb{R}$. The inner product of two functions $f, f^{\prime} \in \mathcal{H}$ could be defined as $\langle f, f^{\prime} \rangle_{\mathcal{H}}= \int_{\mathbf{z} \in\mathcal{Z}} f(\mathbf{z})\cdot f^{\prime}(\mathbf{z})d\mathbf{z}$. A {\it Reproducing Kernel Hilbert Space} (RKHS) is a special Hilbert space of functions. It has two properties:
\begin{enumerate}
    \item For any $\mathbf{z}\in\mathcal{Z}$, there is a function $f = \phi(\mathbf{z}) \in \mathcal{H}$, where $\phi: \mathcal{Z}\longrightarrow\mathcal{H}$ maps every point $\mathbf{z}$ in the space $\mathcal{Z}$ to a function $f$ in the space $\mathcal{H}$.
    For all $f\in\mathcal{H}$ and $\mathbf{z}\in\mathcal{Z}$, $f(\mathbf{z}) = \langle f, \phi(\mathbf{z}) \rangle_{\mathcal{H}}$, implying that the evaluation of any function $f \in \mathcal{H}$ at any point $\mathbf{z} \in \mathcal{Z}$ is equal to the inner product of $f$ and $\phi(\mathbf{z})$ defined in the space $\mathcal{H}$.
\end{enumerate}
When letting $f = \phi(\mathbf{z}^{\prime})$, i.e., when $f$ is a feature map of another data point $\mathbf{z}^{\prime}$, we obtain $f(\mathbf{z}) = \phi(\mathbf{z}^{\prime})(\mathbf{z}) = \langle \phi(\mathbf{z}^{\prime}), \phi(\mathbf{z}) \rangle_{\mathcal{H}}$. Suppose we let $\kappa: \mathcal{Z} \times \mathcal{Z} \longrightarrow \mathbb{R}$ be a two variable function defined by
\begin{equation}
    \kappa(\mathbf{z}^{\prime}, \mathbf{z}) \equiv \phi(\mathbf{z}^{\prime})(\mathbf{z}),
\end{equation}
the property {\bf (2)} obtains the {\it reproducing property}:
\begin{equation}
    \kappa(\mathbf{z}^{\prime}, \mathbf{z}) = \langle \phi(\mathbf{z}^{\prime}), \phi(\mathbf{z}) \rangle_{\mathcal{H}},
\end{equation}
where $\kappa$ is $\mathcal{H}$'s {\it reproducing kernel}. To summarize, a RKHS $\mathcal{H}$ is fully characterized by the reproducing kernel $k$. (1) It maps a point in $\mathcal{Z}$ to a function in $\mathcal{H}$ through $\phi(\cdot)$; (2) The evaluation of $f$ at $\mathbf{z}$ is equal to the inner product of $f$ and the feature map $\phi(\mathbf{z})$; (3) The kernel of two elements in $\mathcal{Z}$ is equal to the inner product of their feature map functions in the RKHS $\mathcal{H}$.

\subsection{Kernel Mean Embedding and Kernel Distance}\label{appendix:kernel_distance}
The idea of {\it kernel mean embedding} is to extend the feature map $\phi: \mathcal{Z} \longrightarrow \mathcal{H}$ from individual data points $\mathbf{z}$ to the set of probability distributions $\mathcal{P}$ on $\mathcal{Z}$, through $\mu: \mathcal{P} \longrightarrow \mathcal{H}$ \cite{sriperumbudur2010hilbert}:
\begin{equation}
\begin{split}
    \mu_{p} =\int_{\mathcal{Z}}\phi(\mathbf{Z})dp(\mathbf{Z}) \\
    \hat{\mu}_{p} = \frac{1}{N}\sum_i\phi(\mathbf{z}_i),
\end{split}
\end{equation}
where the uppercase $\mathbf{Z}$ represents a random variable. In this sense, we transform the distribution $p\in\mathcal{Z}$ to an element $\mu_p\in\mathcal{H}$, so that the kernel mean representation $\mu_p$ captures all information about the distribution $p$ \cite{fukumizu2004dimensionality,sriperumbudur2010hilbert}. Ideally, the mean mapping $p\mapsto\mu_p$ should be {\it injective}, that is, $\|\mu_p - \mu_q\|_{\mathcal{H}}=0$ if and only if $p=q$. In other words, it should create a one-to-one mapping between distributions and the embeddings so that there is no information loss when mapping distributions into the Hilbert space. Not all RKHS satisfy this criterion, Fukumizu et al. \cite{fukumizu2007kernel} shows that Gaussian and Laplacian are two popular kernels that ensure the RKHS embedding maps all distributions uniquely. The RKHS satisfying this criterion is known as {\it characteristic RKHS}, and their kernel functions are {\it characteristic kernels}. When $\mathcal{H}$ is a special case $\mathcal{H}_1=\{f:\|f\|_{\mathcal{H}}\leq 1\}$, where all the elements are limited by a unit ball, we would obtain a {\it kernel distance} of $p,q$ \cite{gretton2006kernel,gretton2012kernel}:
\begin{equation}
    D_{\mathcal{H}_1}[p,q] = \|\mu_{p} - \mu_{q}\|_{\mathcal{H}_1}^2.
\end{equation}
It has been proved that $\mathcal{H}_1$ meets the criterion that $D_{\mathcal{H}_1}[p,q]=0$ if and only if $p=q$ \cite{gretton2006kernel,gretton2012kernel}.

\subsection{Kernel Distance and Integral Probability Metric (IPM)}
Given two probability measures $p$ and $q$ on a measurable space $\mathcal{Z}$, an {\it integral probability metric} (IPM) is defined as \cite{muller1997integral}:
\begin{equation}
    \gamma_{\mathcal{F}}[p,q] = \mathrm{sup}_{f\in\mathcal{F}}\bigg[\int_{\mathcal{Z}} f(\mathbf{Z})dp(\mathbf{Z}) - \int_{\mathcal{Z}} f(\mathbf{Z}^{\prime}) d q(\mathbf{Z}^{\prime})\bigg],
\label{equation:imp}
\end{equation}
where $\mathcal{F}$ is a space of real-valued bounded measurable functions on $\mathcal{Z}$. The function class $\mathcal{F}$ fully characterizes the IPM. There is obviously a trade-off on the choice of $\mathcal{F}$. That is, on one hand, the function class must be rich enough so that $\gamma_{\mathcal{F}}[p,q]$ vanishes if and only if $p = q$. On the other hand, the larger the function class $\mathcal{F}$, the more difficult it is to estimate $\gamma_{\mathcal{F}}[p,q]$. Thus, $\mathcal{F}$ should be restrictive enough for the empirical estimate to converge quickly \cite{sriperumbudur2012empirical}.

When letting $\mathcal{F}$ be a unit ball in a RKHS, i.e., $\mathcal{F} = \mathcal{H}_1 :=\{f: \|f\|_{\mathcal{H}}\leq 1\}$, and its feature map is $\phi$, the resulting metric would become a {\it kernel distance} \cite{gretton2006kernel,gretton2012kernel}:
\begin{equation}
\begin{split}
    \gamma_{\mathcal{H}}[p,q] &= \mathrm{sup}_{f\in\mathcal{H}_1}\bigg[ \langle f, \int_{\mathcal{Z}} \phi(\mathbf{Z})dp(\mathbf{Z}) - \int_{\mathcal{Z}} \phi(\mathbf{Z}^{\prime})dq(\mathbf{Z}^{\prime}) \rangle \bigg] \dashrightarrow \mathrm{reproducing \: property}\\
    &= \bigg\| \int_{\mathcal{Z}} \phi(\mathbf{Z})dp(\mathbf{Z}) - \int_{\mathcal{Z}} \phi(\mathbf{Z}^{\prime})dq(\mathbf{Z}^{\prime}) \bigg\|_{\mathcal{H}_1}^2 \dashrightarrow \mathrm{Cauchy}-\mathrm{Schwarz \: inequality} \\
    &= \| \mathbf{\mu}_p - \mathbf{\mu}_q \|_{\mathcal{H}_1}^2 \triangleq D_{\mathcal{H}_1}[p,q].
\end{split}
\end{equation}
Therefore, kernel distance is a special instance of IPM \cite{sriperumbudur2010hilbert}.

% The choice of $\mathcal{F}$ is critical: $\mathcal{F}$ needs to be (1) ``rich'' enough to satisfy the one-to-one mapping between distributions and embeddings; Besides, it needs to be (2) ``restrictive'' enough for the empirical estimate to converge quickly to its expectation as the sample size or feature dimensions increase.

A number of covariate imbalance metrics, such as standardized differences of means, higher-order moments and interactions, the overlapping coefficient, as well as the Kolmogorov-Smirnov distance, are proposed in causal inference literatures \cite{austin2015moving,austin2009balance}. A natural question is: why choosing kernel distance? As seen in Equation~(\ref{equation:imp}), different options of $\mathcal{F}$ leads to different metrics, $\gamma_{\mathcal{F}}[p,q]$ is the total variation metric when $\mathcal{F} = \{f: \|f\|_{\infty} \leq 1\}$ while it is the Kolmogorov distance when $\mathcal{F} = \{\mathbf{1}_{(-\infty,t]:t\in\mathbb{R}^d}\}$ \cite{sriperumbudur2010hilbert,sriperumbudur2012empirical}. Kernel distance is noted to have a number of important advantages compared with the other metrics regarding these two aspects.
\begin{enumerate}
    \item Kernel distance is dependent only on the kernel, and kernels can be defined on arbitrary domains in a flexible manner. It has been proved that the choice of $\mathcal{F} = \mathcal{H}_1$ satisfies the ``injective'' one-to-one mapping criterion and maps each distribution into an unique embedding in $\mathcal{H}_1$. Instead, standardized differences of means, higher-order moments and interactions are basically restricting $\mathcal{F}$ to be simple polynomial functions, failing to meet the injective mapping criterion.
    \item It is computationally cheaper. The empirical estimate converges at a faster rate, and the rate of convergence is independent of the dimension of the space. In contrast, other metrics such as Kullback-Leibler divergence exhibits arbitrarily slow rates of convergence depending on the distributions \cite{sriperumbudur2010hilbert}.
\end{enumerate}

\subsection{Three Estimators of Expected Outcome}\label{appendix:welfare_estimators}

The central hurdle in learning policies from observational data arises from the partial observability of outcomes. In each round, only the outcome corresponding to the chosen action is revealed, leaving the investigators unaware of the expected outcome $W(\pi)$ and complicating the optimization task $\mathrm{argmax}_{\pi\in\Pi}W(\pi)$. To address this issue, one needs to {\it estimate} the expected outcome of a given policy. This section reviews three estimators of $W(\pi)$: {\it direct method}, {\it inverse propensity score} estimator \cite{lambert2007more,horvitz1952generalization}, as well as {\it doubly robust} estimator \cite{dudik2014doubly}, and discusses their pros and cons \cite{beygelzimer2009offset,dudik2014doubly}.

The first one is called {\it direct method} (DM). DM directly learns two response models $\widehat{m}(0,\mathbf{Z}), \widehat{m}(1,\mathbf{Z}):\mathbf{Z}\mapsto R$ based on data samples of the control versus the treated subsets, and use the estimated functions to directly compute outcomes, written as
\begin{equation}
    \widehat{W}^{\mathrm{DM}}(\pi) = \frac{1}{N}\sum_{i=1}^{N}\bigg[ \sum_{a\in\{0,1\}}\pi(a|\mathbf{Z}_i)\widehat{m}(a,\mathbf{Z}_{i})\bigg]=\frac{1}{N}\sum_{i=1}^{N}\bigg[ \pi(1|\mathbf{Z}_i)\widehat{m}(1,\mathbf{Z}_{i}) + \pi(0|\mathbf{Z}_i)\widehat{m}(0,\mathbf{Z}_{i})\bigg].
\end{equation}
% A problem with this method is that the estimator is formed without the knowledge of $e$, and hence when we approximate a data sample's counterfactual outcome by switching the treatment from $X_i$ to $1-X_i$, we might wrongly extrapolating from other samples with similar features.
The second kind, called {\it inverse propensity score} (IPS) \cite{horvitz1952generalization}, learns $\widehat{e}: \mathbf{Z}\mapsto [0,1]$ that compute a score of being treated given a vector input. Then it uses importance weighting to correct for the incorrect proportions of treatments in the historic data,

% \begin{equation}
%     \widehat{W}^{\mathrm{IPS}}(\pi) = \frac{1}{N}\sum_{i=1}^{N}\bigg[ \frac{\mathbf{1}(\pi(\mathbf{Z}_i) = X_i)}{\widehat{e}(\pi(\mathbf{Z}_i)|\mathbf{Z}_{i})} \cdot Y_{i} \bigg].
% \end{equation}

\begin{equation}
    \widehat{W}^{\mathrm{IPS}}(\pi) = \frac{1}{N}\sum_{i=1}^{N}\bigg[ \frac{\pi(X_i|\mathbf{Z_i})}{\widehat{e}(X_i|\mathbf{Z}_{i})} \cdot Y_{i} \bigg].
\end{equation}

Both DM and IPS have drawbacks. In DM, the fitted outcome function $\widehat{m}$ might give a very different response against the true one when altering treatment actions, especially for the cases where the true outcome generations for treatment 0 and 1 are very different \cite{beygelzimer2009offset}. In IPS, the fitted propensity score $\widehat{e}$ might be very different from the unknown ground truth. IPS typically has a much larger variance and gets more severe when the denominator $\widehat{e}(X_i|\mathbf{Z}_{i})$ gets smaller in high probability areas under $\pi$ \cite{dudik2014doubly}.

The third one is {\it doubly robust} estimator (DR) \cite{robins1995semiparametric,dudik2014doubly,cassel1976some}. Doubly robust estimators are a class of statistical estimators commonly used in causal inference and observational studies. They are designed to provide consistent and unbiased estimates of treatment effects even when either the response model $m$ or the propensity score model $e$ is misspecified. We used the most common augmented inverse probability weighting (AIPW) estimator proposed by \cite{dudik2014doubly},
\begin{equation}
\begin{split}
\widehat{W}^{\mathrm{DR}}(\pi) &= \frac{1}{N}\sum_{i} \bigg[ \sum_{a\in\{0,1\}}\pi(a|\mathbf{Z}_i)\widehat{m}(a, \mathbf{Z}_{i}) + \frac{\pi(X_i|\mathbf{Z}_{i})}{\widehat{e}(X_i|\mathbf{Z}_{i})} \cdot \bigg(Y_{i} - \widehat{m}(X_i, \mathbf{Z}_i) \bigg) \bigg] \\
&=\begin{cases}
 \text{$\frac{1}{N}\sum_{i} [ \widehat{m}(0, \mathbf{Z}_{i}) + \frac{\delta(X_i=0)}{\widehat{e}(X_i|\mathbf{Z}_{i})} \cdot (Y_{i} - \widehat{m}(X_i, \mathbf{Z}_i) ) ]$}, & \text{if $\pi(0|\mathbf{Z}_i)=1$} \\
 \text{$\frac{1}{N}\sum_{i} [ \widehat{m}(1, \mathbf{Z}_{i}) + \frac{\delta(X_1=1)}{\widehat{e}(X_i|\mathbf{Z}_{i})} \cdot (Y_{i} - \widehat{m}(X_i, \mathbf{Z}_i) ) ]$}, & \text{if $\pi(1|\mathbf{Z}_i)=1$}
\end{cases}
% \widehat{W}^{\mathrm{DR}}(\pi) &= \frac{1}{N}\sum_{i} \bigg[ \widehat{m}(\pi(\mathbf{Z}_{i}), \mathbf{Z}_{i}) + \frac{\mathbf{1}(X_i=\pi(\mathbf{Z}_i))}{\widehat{e}(\pi(\mathbf{Z}_i)|\mathbf{Z}_{i})} \cdot \bigg(Y_{i} - \widehat{m}(X_i, \mathbf{Z}_i) \bigg) \bigg].
\end{split}
\label{eq:drestimator}
\end{equation}
It combines the inverse probability weights from the propensity score model with the residuals from the response model. By incorporating both models, doubly robust estimators gives a consistent and unbiased estimate even if $\widehat{e}$ or $\widehat{m}$ is incorrectly misspecified. \revision{We provide a simple example to showcase this property. Imagine a scenario where male $Z=1$ and female participants $Z=0$ would receive a treatment with a probability $e(Z)=0.6Z+0.4$. Hence 6 out of 10 men are treated while 4 out of 10 women are treated. The ground truth response model is written as $m(X,Z)=5X+10(Z+1)$. It means that the untreated outcome for men is 20 and 10 for women, and the treatment effect be 5 for both subgroups. However, $m$ or $e$ might be misspecified without knowing its true parametric forms. Let's assume that the estimator $\widehat{e}=e$ but $\widehat{m}$ is misspecified as $\widehat{m}(X,Z)=m(X,Z)+3$. It means the estimated treated/control outcomes for men are 28 vs 23, the values for women are 18 vs 13. If we assume the target policy is $\pi(1|\mathbf{Z})=1$ requiring all participants to be treated, the true average outcome would be $W(\pi)=\frac{1}{20}[25\cdot 10 + 15 \cdot 10]=20$. Based on Equation~(\ref{eq:drestimator}), $\widehat{W}^{\mathrm{DR}}(\pi)=\frac{1}{20}[6 \cdot (28 + \frac{1}{0.6}(25 - 28)) + 4\cdot 28 + 4 \cdot (18 + \frac{1}{0.4}(15 - 18)) + 6 \cdot 18]=20$, $\widehat{W}=W$ even though $\widehat{m}$ is misspecified. Alternatively, let's assume $\widehat{m}$ to be correct and $\widehat{e}$ misspecified, the second term $Y_i-\widehat{m}(X_i,\mathbf{Z}_i)$ in Equation~(\ref{eq:drestimator}) vanishes to zero and the first term $\widehat{m}(1,\mathbf{Z}_i)$ ensures an accurate estimate of the expected outcome, making the estimator still valid.} For theoretical and empirical results regarding bias and variance of the DR estimator as a function of errors in $m$ and $e$, please refer to \cite{dudik2014doubly}.

\subsection{{\it C}-fold Cross-Fitted Approach to Estimate Propensity Score Model and Response Model}\label{appendix:cross_fitting}
{We divide data into $C$ folds, and then estimate each pair of propensity score model and response model, $\widehat{m}^{-c}, \widehat{e}^{-c}$, only using the $C-1$ folds of data except for the $c$-th fold, leading to $C$ pairs of two models in total. Suppose for a certain data unit $i$ which belongs to the data fold $c$, when computing its expected outcome $\widehat{W}^{\mathrm{DR}}_i(\pi)$ (see Equation~\ref{eq:drestimator}), we particularly use the pair of models $\widehat{m}^{-c}, \widehat{e}^{-c}$ that do not include data fold $c$. Cross-fitting is a commonly used technique in statistical estimation \cite{chernozhukov2018double} to reduce overfit. In this work, we have used logistic regression to estimate $\widehat{e}$ and LASSO linear regression model to estimate $\widehat{m}$, and set $C$ to be 3.}

\section{Supplementary Information on \deparadoxtree Experiments}\label{appendix:appendix_hybridcompr}

Figure~\ref{fig:appendix_hybridcompr} depicts two trees generated by the \deparadoxtree and the baseline method. When both flip probabilities are set to 10\%, both methods face challenges in selecting ground-truth variables for splitting. The proposed method misses two confounders but successfully identifies {\tt p2004} as a feature to uncover nested mixed effects. Overall, the \deparadoxtree maintains a shallow structure. In contrast, the baseline method produces seven balanced terminals by splitting on five features: {\tt sex}, {\tt p2000}, {\tt hh\_size}, {\tt g2002}, and {\tt p2002}. However, four of these split features are incorrect. The tree structure becomes more complex and deeper. Its split criterion aims to model propensity score, potentially leading to a deep tree proficient in predicting treatment but failing to reveal true subgroup-level covariate imbalances.

\begin{figure}[H]
    \centering
    \includegraphics[width=\linewidth]{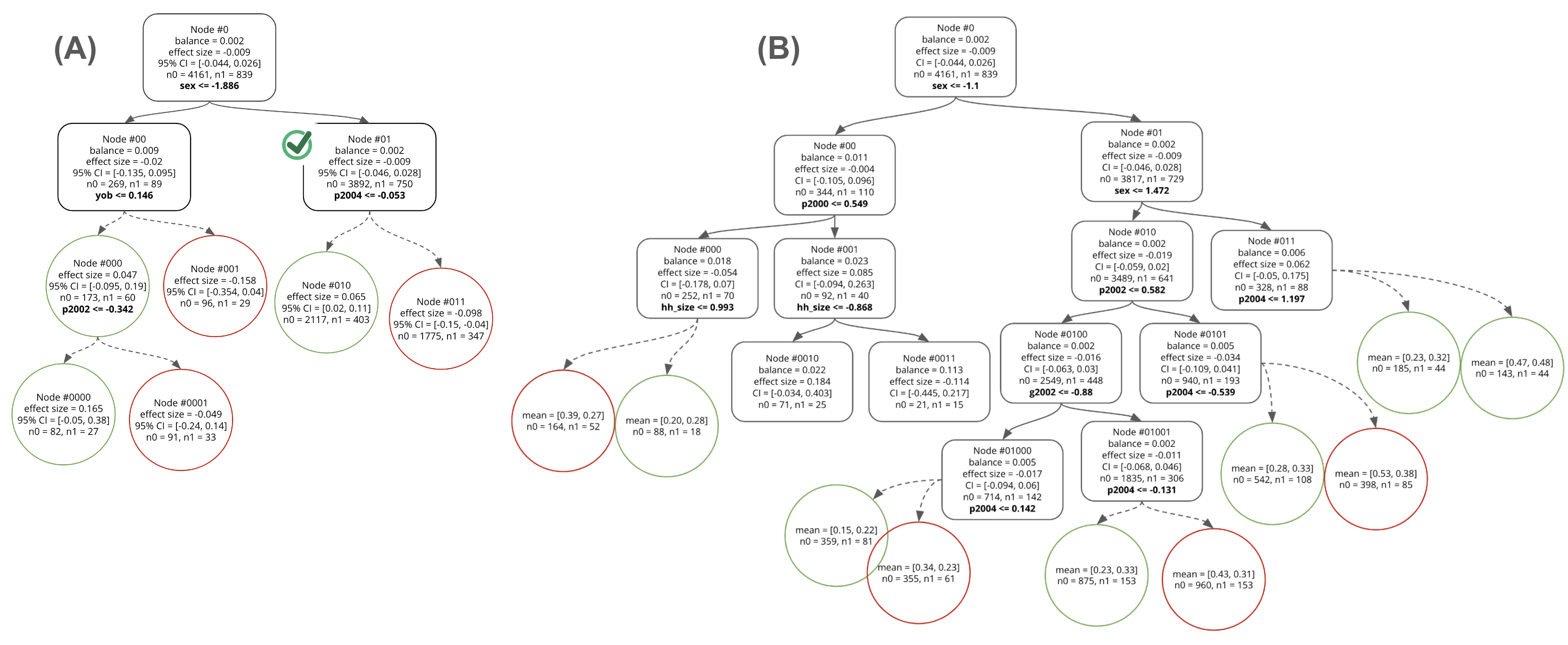}
    \caption[Comparing \deparadoxtree and baseline method on the hybrid voting data when flip probabilities are set to be 10\%.] {Comparing trees from our \deparadoxtree method (A) and baseline (B) on the hybrid voting data, when $p_{X_{\mathrm{flip}}} = 0.1$ and $p_{Y_{\mathrm{flip}}} = 0.1$. Node boxes include ID, split criterion, kernel balance, effect size with 95\% CI, and sample size for treated (NT) and control (NC) groups. Square nodes are to mitigate confounding bias, circular nodes are to search mixed effects. Marked nodes match the ground-truth variable for subgroup specification. Same tree configurations are used for both \deparadoxtree and baseline $n_{\mathrm{min}} = 30, n^1_{\mathrm{min}} = n^0_{\mathrm{min}} = 15, d_1 = 4, d_2 = 2$.}
    \label{fig:appendix_hybridcompr}
\end{figure}
\begin{figure}[h]
    \centering
    \includegraphics[width=0.9\linewidth]{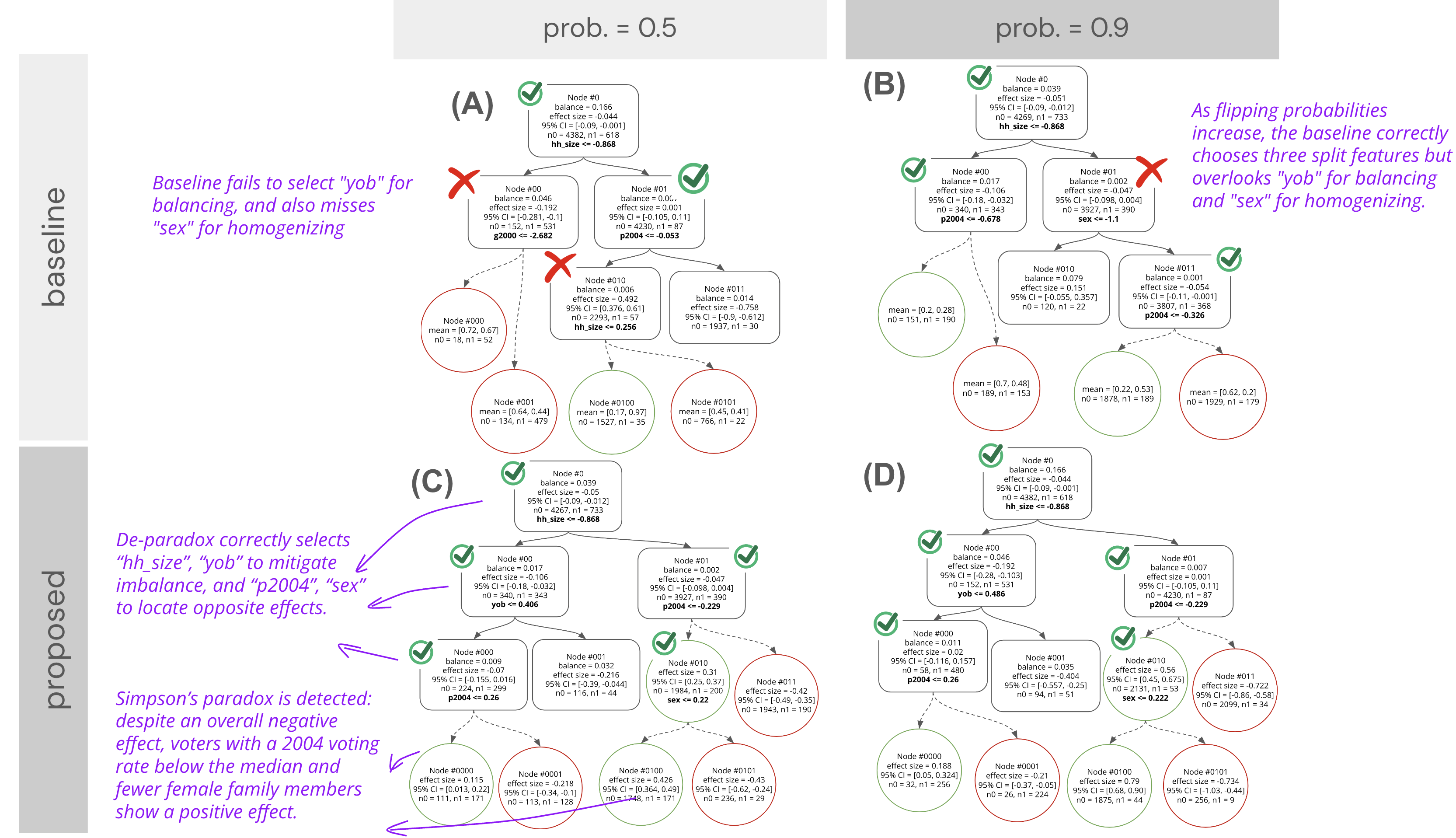}
    \caption{A large version of Figure~\ref{fig:hybridcompr}. Comparing \deparadoxtree vs baseline method on the hybrid voting datasets.}
    \label{fig:hybridcompr_original}
\end{figure}
\section{Applying Non-Decision-Tree Method on the Hybrid Voting Dataset}\label{appendix:nontree}
\begin{figure}[H]
    \centering
    \includegraphics[width=\linewidth]{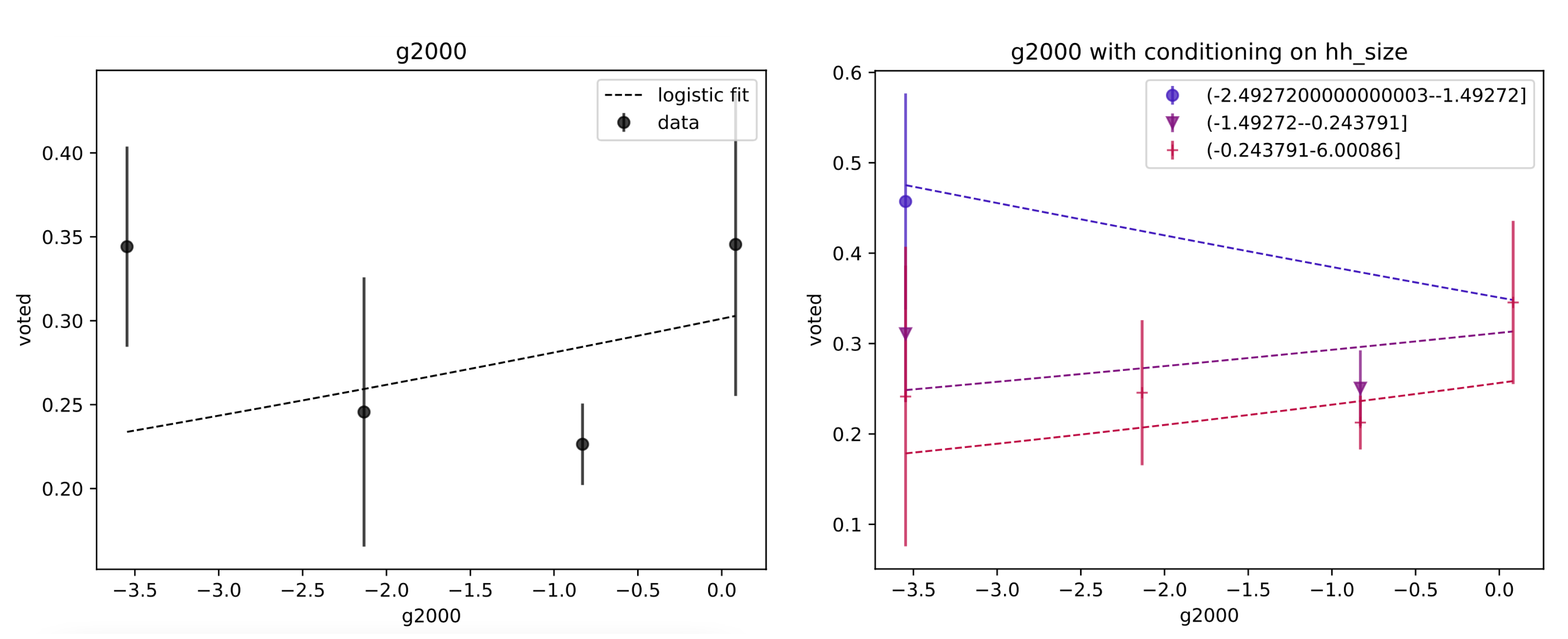}
    \caption[Applying a non-decision-tree baseline \cite{alipourfard2018using} to the hybrid voting dataset $p_{{X_\mathrm{flip}}} = 0.9$ and $p_{{Y_\mathrm{flip}}} = 0.9$.] {\xtengR{Applying a non-decision-tree baseline \cite{alipourfard2018using} to the hybrid voting dataset where $p_{{X_\mathrm{flip}}} = 0.9$ and $p_{{Y_\mathrm{flip}}} = 0.9$. The left plot shows the aggregated trend between {\tt g2000} and outcome {\tt voted}. The right plot shows the disaggregated trends detected by the method conditioning on {\tt hh\_size}.}}
    \label{fig:appendix_nontree}
\end{figure}
\xtengR{We ran one of the methods ~\cite{alipourfard2018using}\footnote{Source code: https://github.com/ninoch/Trend-Simpsons-Paradox} on our hybrid voting dataset where $p_{{X_\mathrm{flip}}} = 0.9$ and $p_{{Y_\mathrm{flip}}} = 0.9$ (see Section~\ref{sec:results::hybrid}). This method permutes variables to identify bivariate ``Simpson’s pairs'' where an aggregate trend reverses when conditioning on another variable. Figure \ref{fig:appendix_nontree} shows that it found one and only one Simpson's pair, ({\tt g2000}, {\tt hh\_size}), suggesting the correlation between {\tt g2000} (household voting rate in 2000) and the outcome {\tt voted} (whether a person voted in the August 2006 primary election) reverses when conditioning on {\tt hh\_size} (household size). However, it failed to reveal any injected patterns---injected imbalance into {\it year of birth} and {\it household size}, and injected heterogeneous effects into {\it p2004} and {\it sex}---behind the Simpson's paradox. This method simply enumerates bivariate trend reversals without supporting multi-covariate disaggregation or addressing causal mechanisms.}

\end{document}